\documentclass[runningheads]{llncs}

\usepackage{eccv}

\usepackage{eccvabbrv}

\usepackage{graphicx}
\usepackage{booktabs}
\usepackage{tabularx}
\usepackage{float}
\usepackage[dvipsnames]{xcolor}
\usepackage{bm}
\usepackage[percent]{overpic}
\usepackage{makecell}
\usepackage{xspace}
\usepackage{wrapfig}

\usepackage[accsupp]{axessibility}  % Improves PDF readability for those with disabilities.

\usepackage{hyperref}

\def\shortname{TLControl\xspace}

\newcommand{\highlight}[1]{{#1}}
\newcommand{\ZY}[1]{{\color{black}#1}}
\newcommand{\first}[1]{\textbf{{#1}}}

\usepackage{orcidlink}

\begin{document}

\title{TLControl: Trajectory and Language Control for Human Motion Synthesis} 

\titlerunning{TLControl: Trajectory and Language Control for Human Motion Synthesis}

\author{Weilin Wan\inst{1, 2}\orcidlink{0000-0002-0555-8295} \and
Zhiyang Dou\inst{1, 2, 4}\orcidlink{0000-0002-0555-8295}  \and Taku Komura\inst{1}\orcidlink{0000-0002-2729-5860} 
\and Wenping Wang\inst{3}\orcidlink{0000-0002-2284-3952} \and \\ Dinesh Jayaraman\inst{2 \dag}\orcidlink{0000-0002-6888-3095} \and Lingjie Liu\inst{2 \dag}\orcidlink{0000-0003-4301-1474}}

\authorrunning{W. Wan et al.}

\institute{The University of Hong Kong
\and
University of Pennsylvania
\and
Texas A\&M University
\and
TransGP
\\
\footnotesize \textit{$^{\dag}$denotes equal contributions}
}

\maketitle

\begin{abstract}

Controllable human motion synthesis is essential for applications in AR/VR, gaming and embodied AI. Existing methods often focus solely on either language or full trajectory control, lacking precision in synthesizing motions aligned with user-specified trajectories, especially for multi-joint control. 
To address these issues, we present \shortname, a novel method for realistic human motion synthesis, incorporating both low-level \textbf{T}rajectory and high-level \textbf{L}anguage semantics controls, through the integration of neural-based and optimization-based techniques. Specifically, we begin with training a VQ-VAE for a compact and well-structured latent motion space organized by body parts. We then propose a Masked Trajectories Transformer~(MTT) for predicting a motion distribution conditioned on language and trajectory. Once trained, we use MTT to sample initial motion predictions given user-specified partial trajectories and text descriptions as conditioning. Finally, we introduce a test-time optimization to refine these coarse predictions for precise trajectory control, which offers flexibility by allowing users to specify various optimization goals and ensures high runtime efficiency. Comprehensive experiments show that TLControl significantly outperforms the state-of-the-art in trajectory accuracy and time efficiency, making it practical for interactive and high-quality animation generation.
%\vspace{-2mm}
\keywords{Motion generation \and Text-to-motion \and Trajectory-to-motion}
%\vspace{-2mm}

\end{abstract}

\begin{figure}[!t]
%\vspace{-4.5mm}
    \centering
    \includegraphics[width=0.9\linewidth]{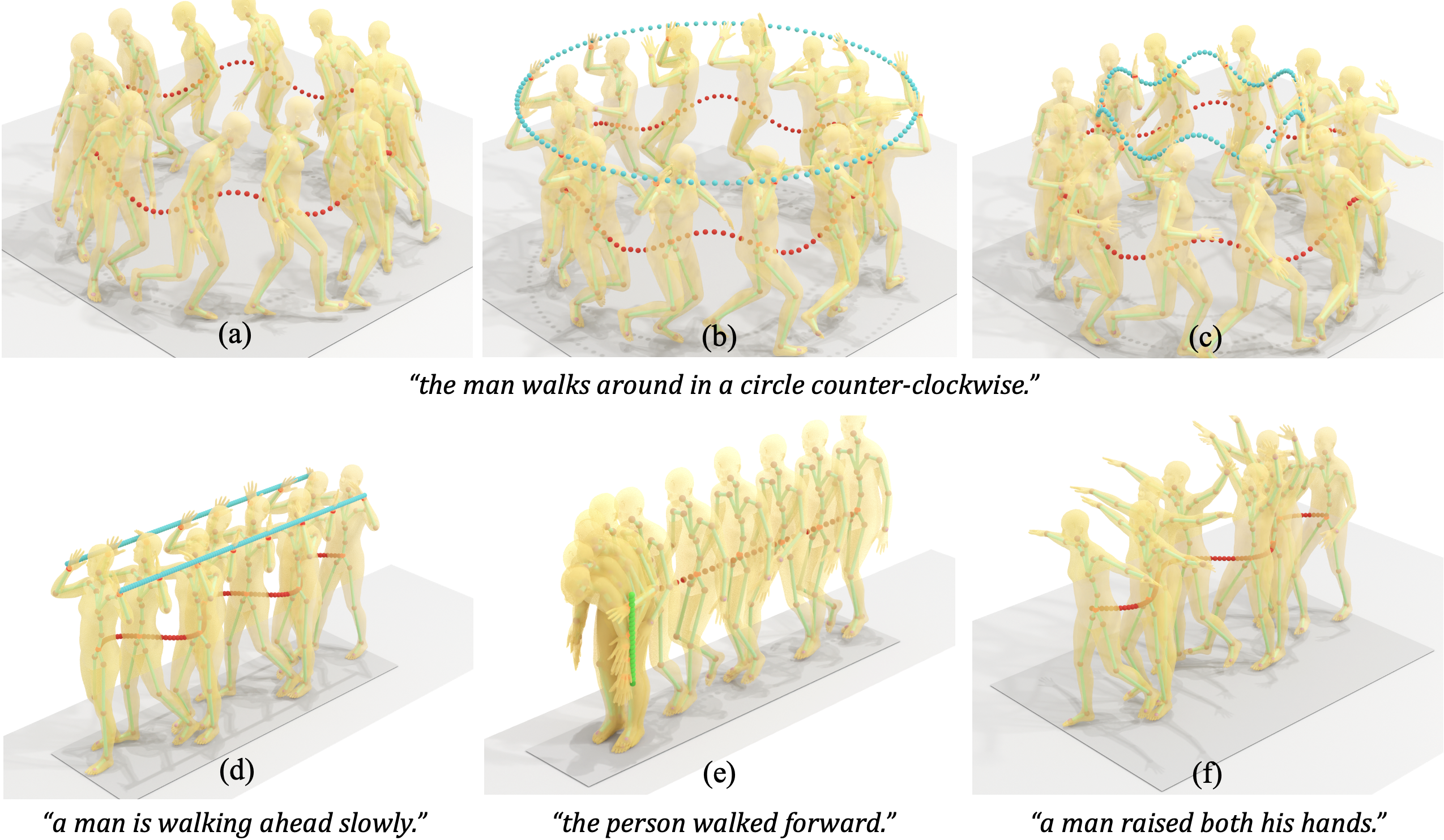}
    %\includegraphics[scale=0.99]{imgs/fig_teaser_motion}
    %\vspace{-2.5mm}
    \caption{\textbf{\shortname}, a novel method for \textbf{Tra}jectory and \textbf{Lan}guage \textbf{Control} for Human Motion Synthesis. The corresponding control joints are highlighted in orange. Our method demonstrates versatile multi-joint controls (see Figures 1a to 1c), the ability to handle complex trajectories (see Figure 1d), multi-stage control (see Figure 1e), and the preservation of language semantics while utilizing trajectory controls (see Figure 1f). The dotted lines represent the input control trajectories defined by users through specifying the parameters of analytical shapes. Note the input trajectories can also be hand drawings from users, or parameters from environment settings (See Fig.~\ref{fig:qualitative_2}). We highly encourage readers to view our supplementary video to see our results.}%\ZY{need to mention the result using trajectory drawn by the users.} %    \label{fig:qualitative_2}
    \label{fig:teaser}
    %\vspace{-9mm}
\end{figure}

% \begin{figure}
% \centering
% \begin{overpic}
% [width=\linewidth]{imgs/fig_teaser_motion} 
% \end{overpic}
% \vspace{-5mm}
% \caption{}
% \label{fig:teasar}
% \end{figure}

\section{Introduction}
\label{sec:intro}
% human motion generation is popular; many works are there.
% but the controllability is still challenging.
Human motion generation is a fundamental problem in computer vision, robotics and graphics, with diverse applications in augmented and virtual reality, the gaming industry, and embodied. In these applications, it is crucial to support user control at different levels on motion generation. 
For example, users may want to have high-level semantics control (e.g., text descriptions) for motion synthesis, and, meanwhile, expect flexible and precise spatial control to ``drag" any joint on arbitrary trajectories. 
% in embodied AI, to generate diverse data,\jd{does this mean diverse data during exploration? since it's not super clear how this would work, maybe warrants a citation or two} we need not only high-level command (e.g., ``turn left and go straight to the room") to control an agent but also low-level control to ``drag" a hand to interact with objects. 
In this work, we study the problem of realistic and precise human motion generation with both high-level language and low-level trajectory controls; \ZY{See an overview in Figure~\ref{fig:teaser}.}

%\cite{Guo_2022_CVPR_t2m, shi2023controllable, petrovich22temos, }
Recently, great advancements have been made in language-conditioned human motion generation~\cite{Guo_2022_CVPR_t2m, shi2023controllable, petrovich22temos, tm2t, ahuja2019language2pose, dou2023c, peng2022ase, mdm, zhang2022motiondiffuse, kim2022flame, karunratanakul2023guided, newmdm, alexanderson2023listen, cong2024laserhuman}. However, most existing works only support language control~\cite{mdm, MLD, T2M-GPT}. Another line of research explores trajectory-controlled motion generation~\cite{rempe2021humor, shi2023phasemp,bensadoun2022neural}, a.k.a., inverse kinematics,
% \jd{Inverse kinematics (at least to me) usually means the end-effectors are fully specified as inputs, and the outputs are the internal joint configurations. If prior work is restricted to only this (barring OmniControl), then we might want to highlight that we are offering more: full motion generation from \textit{incomplete} end-effector specification.}
 but it lacks high-level language semantics control. Furthermore, inverse kinematics can only infer joint configurations from full (complete) trajectories of end-effectors, while our method generates full motions from partial (incomplete) trajectories. The most related works are PriorMDM\cite{newmdm}, GMD~\cite{karunratanakul2023guided}, and a concurrent work, Omnicontrol~\cite{omnicontrol}, that incorporate spatial control signals in a language-conditioned motion generation model, but neither of them can accurately 
align the motion with specified trajectories, especially when controlling multiple joints simultaneously, as \ZY{evidenced by} Figure~\ref{fig:comparison1} and Figure~\ref{fig:comparison2}. Moreover, their motion generation process is time-consuming~(see Sec.~\ref{sec:exp_runtime}) due to inefficient multi-step sampling in the motion diffusion model, restricting its practicality in real-world applications.

% \ZY{@weilin, please have a double check on this general idea below.}
To address these challenges, we propose \shortname, \ZY{leveraging an integrated approach combining neural-based and optimization-based methods}, for Trajectory and Language-controlled human body motion generation. %\ZY{On the one hand, the neural networks learns a joint conditional distribution of body movement given the language semantics and control trajectory. On the other hand, the optimization enforces accurate joint control.} 
Specifically, we propose to operate in a part-structured latent encoding of human motions, learned with a VQ-VAE exploiting knowledge of the skeletal topology of human body. We then synthesize full-body motions from partial trajectory and language input in two steps: a feed-forward coarse estimation, and a subsequent optimization-based refinement.\highlight{ To efficiently sample the feed-forward coarse estimations, we propose a Masked Trajectory Transformer (MTT) that predicts a distribution conditioned on trajectory controls and text descriptions.} \ZY{Then we present an optimization scheme begins with the MTT solution and then iteratively refines it to align with user-specified trajectories. Compared with previous methods, our framework offers flexible controls by allowing users to specify various optimization goals~(Sec.~\ref{sec:exp_motion}) while enjoying high efficiency compared with SOTA methods~(Sec.~\ref{sec:exp_runtime}).}
\ZY{As a result, \shortname produces high-quality motion that aligns with high-level language semantics while accurately following low-level control trajectories derived from 1) specifying parameters of analytical shapes, 2) hand drawings in 3D space, and 3) environment settings. This versatility highlights the practical applicability of our method in a wide range of scenarios.} We conduct extensive experiments to validate the effectiveness of our approach.  Compared to existing methods, \shortname exhibits superior performance in both trajectory accuracy and time efficiency. The results further illustrate that our method empowers users to interactively generate and modify high-quality animations within a brief runtime, affirming its practicality and effectiveness.

In summary, our technical contributions are as follows:
\begin{enumerate}
\item We introduce a novel method \shortname that generates high-quality and precise human motions with language and partial trajectory-based user specifications \ZY{through the tight collaboration of neural-based and optimization-based methods.}

\item \ZY{Our framework effectively learns a compact human body morphology-aware structured latent space using a VQ-VAE, and a motion distribution conditioned on language and trajectory using a Masked Trajectory Transformer. The proposed optimization scheme offers flexible controls by allowing users to specify various optimization goals while ensuring high runtime efficiency.}

\item \ZY{\shortname demonstrates its superior
control accuracy in producing realistic human motion with remarkable runtime efficiency, compared with SOTAs. Our codes will be publicly available upon the paper's publication.}

% in motion synthesis compared to prior work, at comparable or superior performance levels.
%We propose to learn a part-based motion latent space that bridges the multimodalities of human motion, language, and control trajectories by using a masking learning strategy. 
%\item We present a technique for efficient test-time optimization for accurate trajectory control. 
% We introduce a VQ-VAE framework specifically designed for learning priors at the level of individual body parts for human motion synthesis.
% \item We highlight the effectiveness of optimizing motion control at the learned latent space, showing this approach leads to more accurate and time-efficient control outcomes.
% \item We show that our method achieves state-of-the-art performance in both control accuracy and time efficiency.
\end{enumerate}

\section{Related Work}
\label{sec:related_work}
There have recently been significant advances in human motion generation, given its wide applications across virtual reality~\cite{questsim,questenvsim,neural3points}, game development~\cite{tessler2023calm, starke2019neural, starke2021neural,
 holden2017phase, starke2022deepphase}, human behavior analysis~\cite{liu2022close, guo2023student,zhang2023popularization,  zhang2023close, yang2023analysis,chong2020detection,crawford2022impact} and robotics~\cite{wan2022learn, smith2023learning,li2023robust,yamane2013synthesizing,christen2023learning}. Conditional motion generation incorporates diverse, multi-modal inputs such as text prompts~\cite{Guo_2022_CVPR_t2m, shi2023controllable, petrovich22temos, mdm, tm2t, ahuja2019language2pose, kim2022flame, zhang2023tapmo}, action labels~\cite{petrovich21actor, juravsky2022padl, dou2023c, guo2020action2motion}, partial motion sequences~\cite{duan2021single, harvey2020robust, HOGCN}, control signals~\cite{starke2022deepphase, starke2019neural, 2021-TOG-AMP, dou2023c, zhang2023skinned, chen2024taming}, music~\cite{li2021ai, li2022danceformer, alexanderson2023listen}, and images~\cite{rempe2021humor, chen2022learning}. In this section, we conduct a comprehensive review of the existing literature in the field of motion generation. Our reviews covers the studies ranging from unconditional motion generation to conditional approaches, particularly focusing on language-conditioned motion synthesis, known as action-to-motion and text-to-motion. We also explore the integration of trajectory or path controls in these processes, highlighting how these innovations contribute to advancing the field.

\noindent\textbf{Unconditional Motion Generation} This area of research focuses on the autonomous generation of motion sequences without specific guiding conditions or annotations. Pioneering studies, such as those referenced in~\cite{yan2019convolutional,zhao2020bayesian,zhang2020perpetual, raab2022modi, mdm} have made notable strides in this domain. These works stand out for their unique ability to capture and model the entire motion space, leveraging raw motion data to produce diverse and dynamic motion patterns. These methods exhibit innovation in interpreting and representing complex motion dynamics. As we primarily focus on conditional motion generation, our discussion of unconditional motion generation only serves as a contextual background to our primary focus.

\noindent\textbf{Conditional Motion Generation} 
In the realm of conditional,
tasks, ACTOR~\cite{petrovich21actor} presents a class-agnostic transformer VAE as a baseline. It introduces learnable biases within a transformer VAE to encapsulate action for motion generation. But similar to other works in the field of action-to-motion \cite{guo2020action2motion}, it is limited to cover changes in the conditional inputs. Recent studies in the task text-to-motion \cite{petrovich22temos, ahuja2019language2pose, zhang2022motiondiffuse, mdm, kim2022flame, Guo_2022_CVPR_t2m, TMR, Teach1} has emerged as a principal driver, reshaping research frontiers with its user-friendly nature by allowing nature languages input. The advancements can be categorized into two methodologies: joint-latent models and diffusion models. Joint-latent models like TEMOS~\cite{petrovich22temos} typically employs a motion VAE alongside a text variational encoding. These components are then aligned into a compatible latent space by using divergence loss. This method advances the synthesis of human motion sequences from natural language inputs, representing a significant step forward in the field. However, as observed in the misalignment of Gaussian distributions~\cite{tevet2022motionclip}, there are difficulties remaining in aligning the joint distribution of language and human motions.

Recent advancements in image synthesis have been marked by the successful application of diffusion generative models \cite{sohl2015deep}. Building on this progress, these models have now been extended to the domain of human motion synthesis. This novel application is being applied in recent works such as those presented in \cite{mdm, MLD, motionGPT, kim2022flame, omnicontrol, voas2309best, alexanderson2023listen}. Notably, MotionDiffuse~\cite{zhang2022motiondiffuse} emerges as the first text-based motion diffusion model, offering fine-grained instructions for individual body parts. MDM~\cite{mdm} introduces a motion diffusion model that operates directly on raw motion data, further expanding the scope of natural language controls in human motion generation. This method represents a significant leap forward in terms of enabling intuitive, natural language inputs to motion synthesis. 

%However, these models have limited applicability to raw motion data %\jd{previous sentence says they operate on raw motion data, but next sentence says they have limited applicability to raw motion data?}
%due to noise and temporal consistency redundancy, making them susceptible to outliers. 

\noindent\textbf{Motion Generation with Spatial Constraints} 
Despite these previous efforts, the language description itself is still limited to a coarse control over the motion. To tackle the issue, Shafir et al.~\cite{newmdm} proposed PriorMDM to generate long-sequence human motion and joint control signals. To further support more flexible fine-grained motion generation, GMD~\cite{karunratanakul2023guided} integrates spatial constraints by employing a two-stage diffusion strategy. However, GMD only controls the 2D positions of the pelvis, a limitation that reduces its flexibility in many practical scenarios. Recently, the concurrent work Omnicontrol~\cite{omnicontrol} incorporates flexible spatial control signals over different joints by introducing analytic spatial guidance and realism guidance that ensure the generated motion from the diffusion model can tightly conform to the input control signals. However, Omnicontrol struggles to generate body motion accurately aligned with specified trajectories, as we will illustrate. Furthermore, it exhibits a relatively prolonged motion generation time. In comparison with Omnicontrol, our method excels in producing motion that precisely adheres to the control signal, demonstrating superior accuracy and significantly faster runtime.
\section{Method}

% \ZY{double the notions: vector/ scale.}
We aim to generate full-body motions to match user specifications, which consist of two parts: a text description $\mathbf{L}$, and a partial trajectory $\mathbf{R'}$. $\mathbf{R'}$ has $T$ frames, corresponding to the target motion length. \ZY{Notably, we assume that there is no conflict between language and the trajectory lines, and the trajectories themselves represent motions that are feasible in reality.} In each frame $t$, $\mathbf{R'}$ contains the 3D positions of some subset $m_t<n$ of the $n$ full-body joints, corresponding to the ``control joints'' which the user chose to specify for that frame. For example, the user might choose to specify only the head joints for all frames $t$, or even specify different joints in different frames. The desired final output is the full body motion $\mathbf{J} \in \mathbb{R}^{T \times M}$ of the character, \highlight{where $M$ is the dimension of the single frame pose represented by the joint rotations and positions following}~\cite{Guo_2022_CVPR_humanml3d}.

\begin{figure*}[!t]
%\vspace{-4mm}
  \centering
\includegraphics[width=0.96\linewidth]{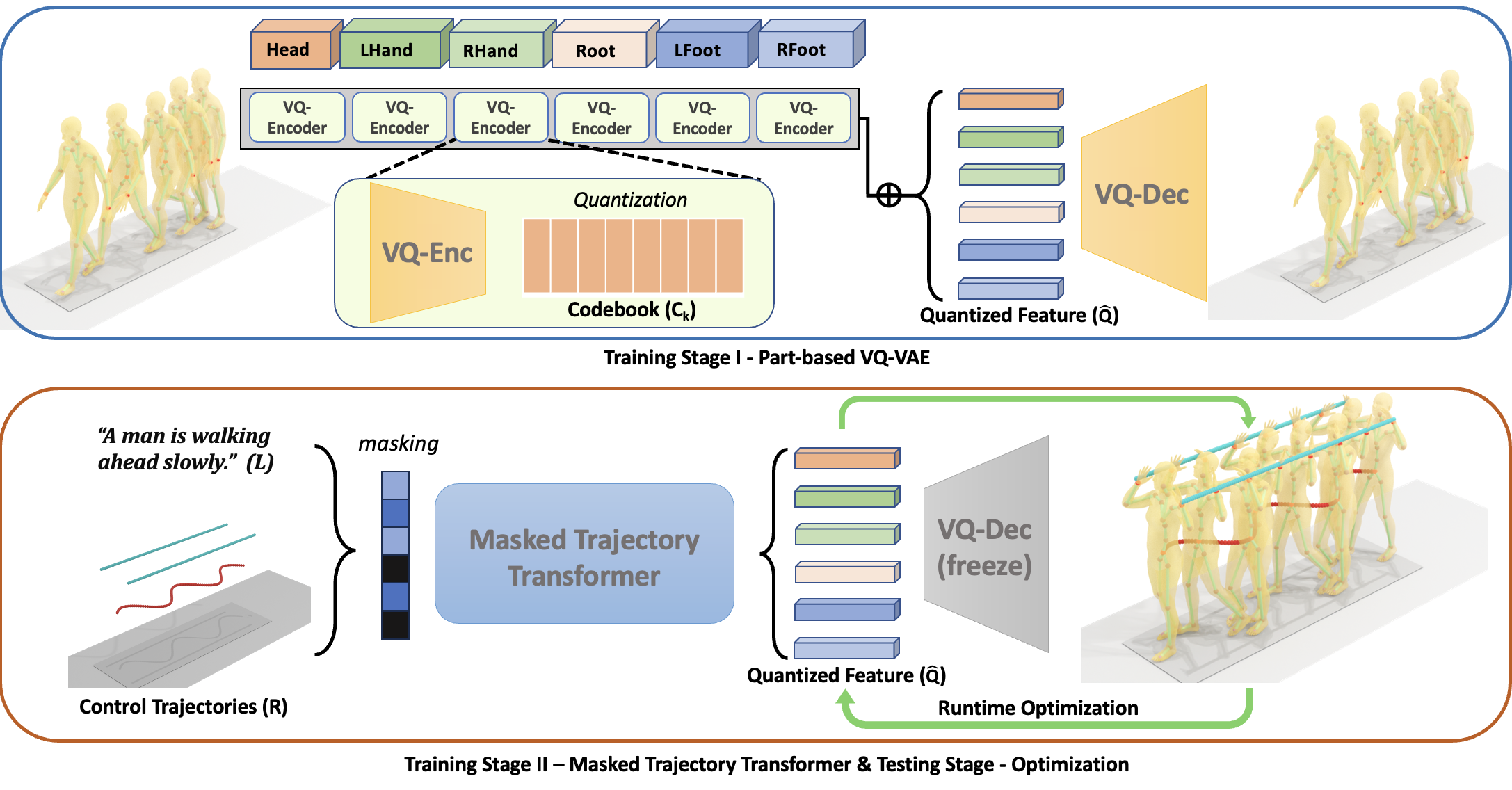}
%\vspace{-4mm}
   \caption{Overview of \shortname framework: At training stage I, we train the part-based VQ-VAE in \ref{subsec:vqvae} for reconstructing human motions. In training stage II, the decoder of the part-based VQ-VAE is frozen and we train the masked trajectory transformer (MTT) in \ref{subsec:transformer} for predicting code indices from control inputs. Finally, at test time, the MTT receives text description and partial control trajectories to predict an initial VQ-VAE quantized code seed, which is refined by run-time optimization as in \ref{subsec:Optimization} before decoding with the VQ-VAE into full body motions.\ZY{remaking.}}
   \label{fig:pipline}
   %\vspace{-7mm}
\end{figure*}

We train \shortname using data in which each sample $(\mathbf{R}, \mathbf{J}, \mathbf{L})$ is a tuple containing a full trajectory specifying some key joints, ground truth full body motions, and language description. 
%Grounding in notation, 
%The goal of TraLanMotion is to generate human motions that match the given text description  and user-specified partial trajectories of key joints faithfully and efficiently. The input to our model consists of the text prompt $\mathbf{L}$ and control trajectories $\mathbf{R}=\{{\mathbf{r}_i}\}_{i=1}^{N}$ where $N$ is the number of key joints used for controlling and each $\mathbf{r}_i \in \mathbb{R}^{H \times 3}$ \jd{3 because 3D coordinates, right? WL: Fixed add 3D} is a 3D trajectory with $H$ being the length. The output is the full body motion of the character. 
To achieve the goal, we first learn a motion embedding by training a VQ-VAE~\cite{VQVAE} to establish part-based latent spaces, representing human body motion at the body part level (Sec.\ref{subsec:vqvae}). 
We then train a text-conditioned Masked Trajectory Transformer (MTT) to output a coarse completion of partially masked trajectories in this learned part-based embedding (Sec.\ref{subsec:transformer}). This coarse completion is then further refined by a 
%To arrive at an initial guess for a full-
%effectively capture motion, language, and control trajectories during this embedding process, we introduce a Masked Trajectory Transformer that takes $\mathbf{L}$ and $\mathbf{R}$ as input. The transformer can reconstruct full control trajectories and select codebook entries generated in the first stage, which are subsequently decoded into a full-body motion sequence (Sec.\ref{subsec:transformer}).
%To effectively capture motion, language, and control trajectories during this embedding process, we introduce a Masked Trajectory Transformer that takes $\mathbf{L}$ and $\mathbf{R}$ as input. The transformer can reconstruct full control trajectories and select codebook entries generated in the first stage, which are subsequently decoded into a full-body motion sequence (Sec.\ref{subsec:transformer}).
%Finally, we develop a 
simple yet effective optimization over the learned latent space to minimize the distance between control targets and completed motions,
%the specific parts of the decoded motion, 
leading to accurate and efficient trajectory control (Sec.~\ref{subsec:Optimization}). An overview of our framework is shown in Fig.~\ref{fig:pipline}.

\subsection{Part-based VQ-VAE}
\label{subsec:vqvae}
Learning discrete representations has proven effective in motion synthesis tasks in recent works \cite{tm2t, T2M-GPT}. These methods use VQ-VAE \cite{VQVAE} for discretizing a continuous motion space, facilitating the capturing of complex motion patterns in a structured manner and enhancing the generation of coherent and high-fidelity motion sequences. However, previous works \cite{tm2t, T2M-GPT} encode the full body into the discrete motion space, treating the human body as a holistic entity. Instead, in this paper, we propose to learn a better representation of part-based motion priors, which have demonstrated impressive results for computer character animation \cite{ jang2022motion, pi2023hierarchical, starke2021neural}. % Frank I modified this part as body part level motion has demonstrated its effectiveness.
To do so, we propose a part-based VQ-VAE that disaggregates the motion at the human body-part level for learning a well-structured and compact latent space. 

Specifically, we first divide all the joints into six joint groups. The first five groups contain the end-effectors and related joints: \emph{Head}, \emph{Left arm},
% \jd{Should we call these left arm, right arm, left leg, right leg? That naturally identifies the groups better than "hand" and "foot".} 
\emph{Right arm}, \emph{Left leg}, and \emph{Right leg}.
% \jd{Are these groups fully specified somewhere? Perhaps a figure? Appendix? WL: We can draw a figure in Appendix} 
The sixth group comprises the \emph{Root} joint only. The input to our VQ-VAE is the ground-truth full body motion $\mathbf{J} \in \mathbb{R}^{T \times M}$ of the character.
%, where $T$ represents the length of the motion and $M$ is the dimension of the single frame pose represented by either joint rotations or positions. 
By reorganizing the features 
%in $P$ 
into these six groups, we can arrange the input $\mathbf{J}$ to be: $\mathbf{J} = [\mathbf{J}_{\emph{Head}}, \mathbf{J}_{\emph{Larm}}, \mathbf{J}_{\emph{Rarm}}, \mathbf{J}_{\emph{Lleg}}, \mathbf{J}_{\emph{Rleg}}, \mathbf{J}_{\emph{Root}}]$. 

In our model, we have a unique encoder $\bm{\mathcal{E}}_k$ for each joint group $\mathbf{J}_{k}$ where $k \in \{\emph{Head}, \emph{Lhand}, \emph{Rhand}, \emph{Lfoot}, \emph{Rfoot}, \emph{Root}\}$ to learn a codebook $\mathbf{C}_{k}$ for each joint group separately. Here, $\mathbf{C}_{k} = \{\mathbf{c}_{i, k}\}_{i=1}^{|C|}$ and $\mathbf{c}_{i, k} \in \mathbb{R}^d$, where $|C|$ represents the number of discrete codes within the codebook and $d$ 
%\jd{weird notation, could we use something more standard, like d?}
is the dimension of the code.  %Our encoder also follows the downsampling in \cite{T2M-GPT}. 
When encoding an input motion \(\mathbf{J}_{k}\) into its latent representation \(\mathbf{Q}_k = \bm{\mathcal{E}}_k(\mathbf{J}_k)\), $\mathbf{Q}_k = \{\mathbf{q}_{t, k}\}_{t=1}^{T/s}$ is of a reduced time length, where $T$ is the total number of frames and $s$ is a temporal downsampling factor, 
%\jd{not sure why downsampling }
and $\mathbf{q}_{t, k} \in \mathbb{R}^{d}$ is the feature for each time step $t$.

Subsequently, we quantize $\mathbf{Q}_k$ using $\mathbf{C}_k$. For each time step feature $\mathbf{q}_{t, k}$, quantization is achieved by locating the closest code $c_{i, k}$ in the codebook by
\begin{equation}
    \hat{\mathbf{q}}_{t, k} = \text{arg min}_{\mathbf{c}_{i, k} \in \mathbf{C}_k} \|\mathbf{c}_{i, k} - \mathbf{q}_{t, k}\|_2,
\end{equation}
where the quantized feature $\hat{\mathbf{q}}_{t, k}$ is reassembled in their original time order to form $\hat{\mathbf{Q}}_k$.

 The decoder $\bm{\mathcal{D}}$ of our part-based VQ-VAE is designed to consider the features of all groups comprehensively to reconstruct the full body motion. Hence the input to $\bm{\mathcal{D}}$ is the concatenation $\hat{\mathbf{Q}} = \bigoplus_k\hat{\mathbf{Q}}_k$ of quantized features from all groups. 
 %and we denote this concatenation operation as $\hat{\mathbf{Q}} = \bigoplus_k\hat{\mathbf{Q}}_k$. 
 This enables the decoder to reconstruct the full body motion as
 %from these concatenated features through 
 $\hat{\mathbf{J}} = \bm{\mathcal{D}}(\hat{\mathbf{Q}})$.

For training our part-based VQ-VAE, we adopt standard loss terms including quantization, commitment and reconstruction losses:
\begin{align}
    \mathcal{L} = \Sigma_k (&\beta \| \text{sg}[\hat{\mathbf{Q}}_k] - \mathbf{Q}_k \|_2 \nonumber +\\
    &\| \hat{\mathbf{Q}}_k - \text{sg}[\mathbf{Q}_k] \|_2) + \| \mathbf{J} - \hat{\mathbf{J}} \|_2
\end{align}
where $\beta$ is a balancing term, while $\text{sg[$\cdot$]}$ denotes the stop-gradient operator. Also, suggested by previous works in this field \cite{T2M-GPT}, we apply code reset and the exponential moving average~\cite{ema} to prevent codebook collapse during training.

\subsection{Masked Trajectory Transformer}
\label{subsec:transformer}
%\ZY{Although we have ", the MTT predicts a conditional probability distribution260 260 based on input trajectories and text descriptions,", it is still not clear that our method is generative? Btw, Gumbal Softmax should be included here.}

Having trained this part-based discrete embedding space, we now train a feed-forward transformer network to produce an initial coarse guess $\hat{\mathbf{Q}}_0$ of the full-body motion embedded in this space, given the specification $(\mathbf{R}', \mathbf{L})$.

%Next, we propose a Masked Trajectory Transformer~(MTT) to bridge the information\jd{I don't know what bridging information means.} of quantized latent motion features, text prompts input and control trajectories. 

Specifically, our Masked Trajectory Transformer (MTT) takes CLIP embeddings of the text prompts, as in prior work~\cite{T2M-GPT, mdm}. 
For trajectories, we adopt the same joint partition in Sec.~\ref{subsec:vqvae} and treat the trajectories of the five end-effectors and the root joint as the ground truth ``full'' trajectories $\mathbf{R}$ 
for training.  
Recall that in our application settings,
%real-world scenarios, 
we would like to enable users to
%users often 
provide rough partial sketches of the desired trajectories to control motion generation.  
%sketch rough and intermittent trajectories rather than.  
%smooth and continuous curves. 
To simulate such specifications when training the MTT, we apply two masking strategies to $\mathbf{R}$ to obtain the masked control trajectories $\mathbf{R'}$:

\noindent $\triangleright$ \textbf{Continuous Trajectory Masking.} Rather than treating each waypoint's masking as an independent event, we consider the influence of adjacent points, which simulates the common scenario where a user's drawn trajectory might have breaks or be sketched in segments. Our method involves first determining a proportion of the trajectory points that need to be masked. Then, we randomly select segments of varying lengths to mask, ensuring that the total number of masked points matches the predetermined proportion.

\noindent $\triangleright$ \textbf{Joint-Level Masking.} Since users may not always provide trajectories for all six joints, we include joint-level masking. This simulates scenarios where only a subset of joint trajectories are specified. Specifically, we randomly select a number of joints and mask corresponding trajectory points during training.

%\vspace{0.03in}
These masking strategies enable the network to learn the overall relationship among human motion, joint trajectories, and textual information. This approach strengthens the model's capability to handle real user specifications, particularly when detailed and complete joint trajectories are unavailable. 

%$\bm{\mathcal{T}}$   \bm{\mathcal{T}}(\mathbf{R}', \bm{L})
%\highlight{
%We train the MTT to process the user-provided observations $(\mathbf{R}', \bm{L})$ to sample the initial quantized latent VQ-VAE motion codes $\hat{\mathbf{Q}}_0$. 

%Specifically, the MTT predicts a probability distribution conditioned on input trajectories and text descriptions, and samples indices $I$ accordingly.

\textbf{Sampling} the initial quantized latent $\hat{\mathbf{Q}}_0$ requires training the MTT to process the user-provided observations $(\mathbf{R}', \bm{L})$ and outputs the logits of selecting code indices $I$. We transform the logits of each code into a one-hot decision vector $\mathbf{v}$ using the Gumbel-Softmax~\cite{jang2016categorical}, a method that enables sampling from distributions by integrating Gumbel noise. In other words, the sampling capability allows for generative motion synthesis. Then, each code is selected using $\mathbf{C}_{k} \cdot \mathbf{v}$, and the resultant selected codes are concatenated to form $\hat{\mathbf{Q}}_0$, which can be decoded using the frozen VQ-VAE decoder $\bm{\mathcal{D}}$. Thus, the loss function for training the MTT is:  
\begin{align}
\mathcal{L} = \mathbb{E}_{\hat{I} \sim P(I)}[-\text{log}P(\hat{I} | R', L)]
+ ||\bm{\mathcal{D}}(\hat{\mathbf{Q}}_0) - \mathbf{J}||_2.
\end{align}
\subsection{Test-Time Optimization}
\label{subsec:Optimization}
At test time, the MTT takes a partially specified control trajectory $\mathbf{R'}$ to produce an initial coarse feed-forward prediction $\hat{\mathbf{Q}}_0$ in the quantized latent code space as above. We then 
%\jd{TODO: explain the test time flow. Text + partial control trajectory R’ -> MTT -> $\hat{Q}$ -> optimize to get refined Q’ -> VQ-VAE decoder -> $\hat{J}$}
%After human motion latent space, we achieve trajectory and language control for human motion synthesis by optimization in the latent space. Given the control trajectories $\mathbf{R'}$ and the initial quantized latent features of the full body motion $\hat{\mathbf{Q}} = \bigoplus_{\text{group}}\hat{\mathbf{Q}}_{\text{group}}$, 
%we 
use test-time optimization over the learned latent space to refine this initial coarse prediction.
%generate body motions that align with the control trajectories.
Specifically, we solve:
%aim to find optimized quantized latent features by:
\begin{equation}
    \hat{\mathbf{Q}} = \text{arg min}_{\mathbf{Q}} \Sigma_{j} || \bm{\mathcal{P}_j}(\bm{\mathcal{D}}( \bm{\mathbf{Q}})) - \mathbf{R}'_j ||_2 
\end{equation}
where $j$ indexes the provided control joints, and 
$\bm{\mathcal{P}_j}$ is a projection function that converts the human pose representation into the global joint positions of joint $j$ which are in the same frame as the waypoints in $\mathbf{R'}$. By applying $\bm{\mathcal{D}}(\mathbf{\hat{Q}})$, we are able to achieve the final body motion controlled by the given trajectories. The optimization is performed using search initialized with the coarse MTT output prediction $\hat{\mathbf{Q}}_0$.  %\jd{please check this all for accuracy}

Since our body-part level motion embedding effectively captures the motion distribution in a compact space, this simple yet effective optimization framework produces high-quality human motion with the desired controllability while maintaining semantic coherence with the textual description.  In contrast to the closest prior work OmniControl~\cite{omnicontrol}  
where limited control specifications are coupled with the network during training, this framework allows for flexible controls, such as simultaneously controlling multiple joints at different time steps. Moreover, it is much more computationally efficient compared with SOTA methods, which we investigate in Sec.~\ref{sec:exp_runtime}.

\section{Experiment}
We conduct extensive experiments of our method and compare it with state-of-the-art methods in terms of motion quality, controllability, accuracy, and efficiency. We further investigate the key design choices in ablation studies.

\begin{figure}[!t]
%\vspace{-6mm}
  \centering
   \includegraphics[width=\linewidth]{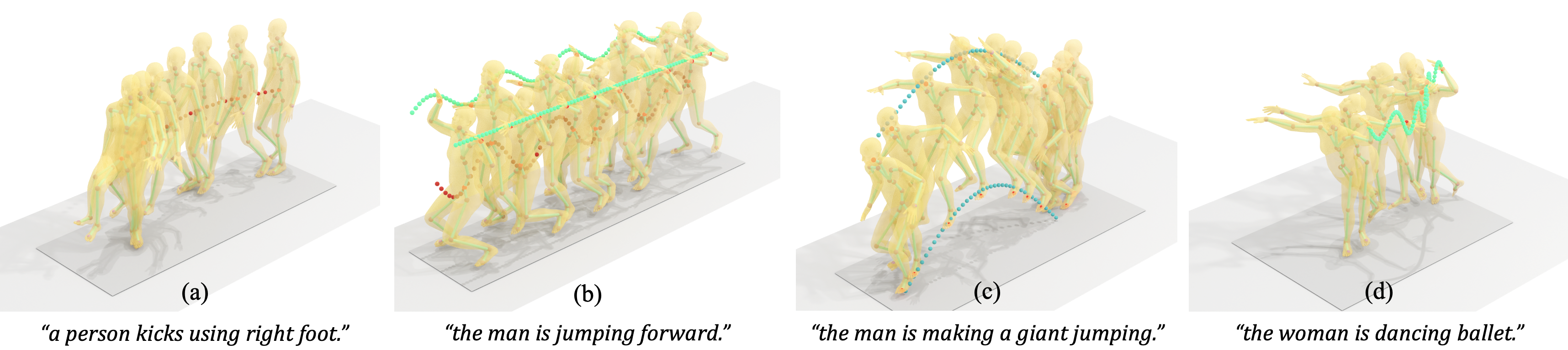}
   %\vspace{-6mm}
   \caption{Qualitative results of \shortname using user-defined trajectories. Figure 3a and Figure 3d demonstrate that our method enables separate controls using language and joint-level trajectories. Figure 3b and Figure 3c showcase the capability of our method to manage multi-joint control simultaneously. Please refer to our supplementary for more qualitative results}
   \label{fig:qualitative}
   %\vspace{-6mm}
\end{figure}

\begin{figure}[!t]
%\vspace{-6mm}
  \centering
   \includegraphics[width=0.95\linewidth]{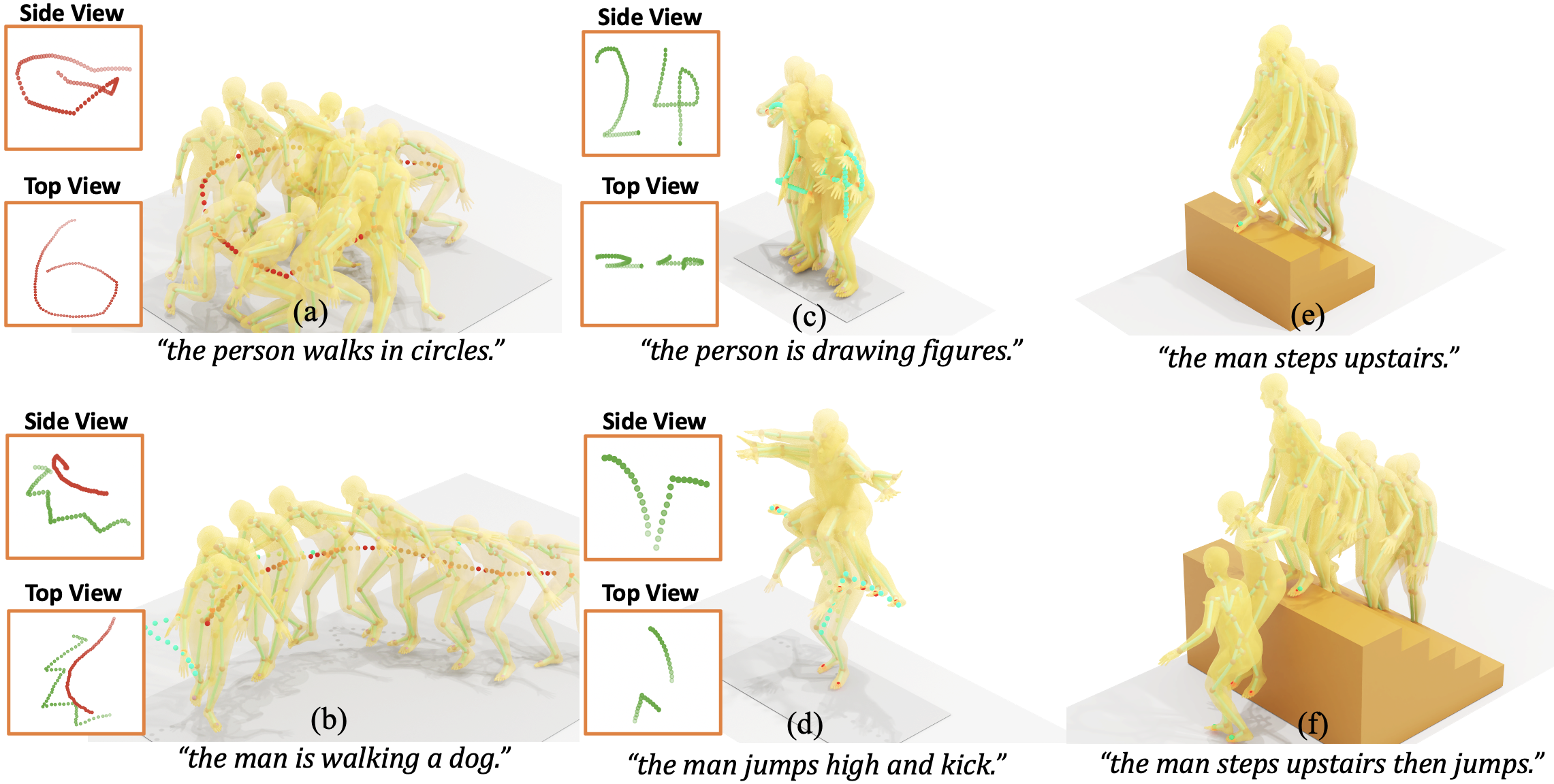}
   %\vspace{-2mm}
   \caption{Figure 4a to 4d: Qualitative results of our method using 3D user hand drawings on the left. Figure 4e and Figure 4f: Showcase that allocating foot placements according to the staircases.%\ZY{the trajectories are not very clear to me.}
   %\ZY{\textbf{language control is too weak}}
   }
   %\vspace{-6mm}
   \label{fig:qualitative_2}
\end{figure}

\subsubsection{Datasets} Our study utilizes two prominent datasets for tasks involving text-based motion generation: KIT Motion Language (KIT-ML) \cite{kitdataset}, and HumanML3D \cite{Guo_2022_CVPR_humanml3d}. The KIT-ML dataset encompasses 3,911 unique human motion sequences and 6,278 individual text annotations, with a frame rate of 12.5 FPS for the motion sequences. On the other hand, HumanML3D offers a more extensive collection of %\jd{is the other dataset not 3D? I assume it is, in which case drop this. Feels odd.}
human motions, featuring 14,616 unique motion capture data paired with 44,970 textual descriptions. For uniformity, all motion sequences in both KIT-ML and HumanML3D are padded to a length of 196 frames.

\subsubsection{Metrics} Following the experiments setting in \cite{omnicontrol}, we evaluate the fidelity of our result using Frechet Inception Distance (FID), R-Precision and Diversity. \textbf{FID} measures the quality of motion created by generative techniques. It calculates the disparity between actual and produced distributions within the feature space of a pre-trained model as being used in \cite{Guo_2022_CVPR_humanml3d}. \textbf{R-Precision} measures a motion sequence against 32 text narratives, one of which is accurate and the remaining 31 are random descriptions. It involves computing the Euclidean distances between the embeddings of the motion and the textual narratives. The accuracy of retrieving the correct text from the motion is then reported based on the top-3 matches. \textbf{Diversity} is evaluated by randomly pairing all generated sequences from the test texts. The average cumulative difference within each pair is then calculated to determine diversity. \textbf{MModality} is the average output variance given the same input.

\begin{figure}
  \centering
  %\vspace{-6mm}
\includegraphics[width=0.93\linewidth]{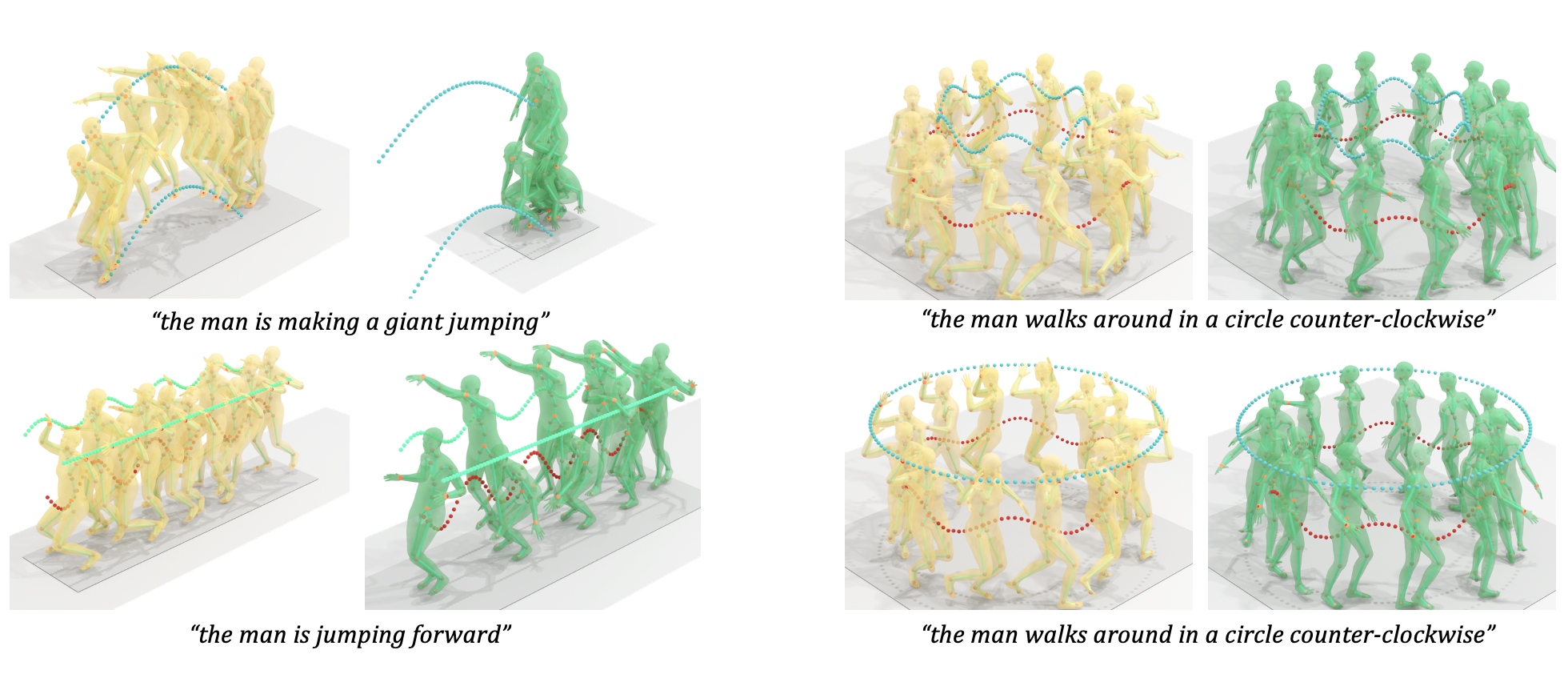}
%\vspace{-2mm}
   \caption{Qualitative comparison results with Omnicontrol\cite{omnicontrol}. Our results are shown in \textcolor{yellow}{Yellow}, while the results of Omnicontrol are depicted in \textcolor{green}{Green}. Please refer to our supplementary video for more details of the comparison. }
   \label{fig:comparison_Omni2}
%\vspace{-6mm}
\end{figure}

We also include an analysis of several metrics to evaluate control accuracy. Following~\cite{omnicontrol, MLD}, we adopt metrics including Trajectory error, Location error, and Average error. These metrics evaluate the 3D control accuracy of joint positions in keyframes when generating motions with trajectory controls. \textbf{Trajectory Error} quantifies the proportion of trajectories that is unsuccessful, characterized by any keyframe's joint location deviating beyond a set threshold. \textbf{Location Error} calculates the percentage of keyframe locations not attained within a specified proximity limit. \textbf{Average Error} is computed as the mean Euclidean distance between the joint positions in the generated motion and the given control trajectories at each keyframe motion step.

\begin{table}[h!]
\centering
\resizebox{0.85\textwidth}{!}{
\begin{tabular}{c|c|c|c|c|c|c|c}
\hline
Method & Control Joint & FID↓ & \makecell[c]{R-precision↑\\(Top-3)} & Diversity→ & \makecell[c]{Traj. Err.↓\\(50 cm, \%)} & \makecell[c]{Loc. Err.↓\\(50 cm, \%)} & \makecell[c]{Avg. Err.\\(cm)↓} \\
\hline
Real & - & 0.002 & 0.797 & 9.503 & 0.00 & 0.00 & 0.00 \\
\hline
MDM &  & 0.698 & 0.602 & 9.197 & 40.22 & 30.76 & 59.59 \\
PriorMDM &  & 0.475 & 0.583 & 9.156 & 34.57 & 21.32 & 44.17 \\
GMD & Pelvis & 0.576 & 0.665 & 9.206 & 9.31 & 3.21 & 14.39 \\
OmniControl &  & \first{0.218} & 0.687 & 9.422 & 3.87 & 0.96 & 3.38 \\
\textbf{Ours} &  & 0.271 & \first{0.779} & \first{9.569} & \first{0.00} & \first{0.00} & \first{1.08} \\
\hline
OmniControl & Head & 0.335 & 0.696 & \first{9.480} & 4.22 & 0.79 & 3.49 \\
\textbf{Ours} &  & \first{0.279} & \first{0.778} & 9.606 & \first{0.00} & \first{0.00} & \first{1.10} \\
\hline
OmniControl & Left Hand & 0.304 & 0.680 & \first{9.436} & 8.01 & 1.34 & 5.29 \\
\textbf{Ours} &  & \first{0.135} & \first{0.789} & 9.757 & \first{0.00} & \first{0.00} & \first{1.08} \\
\hline
OmniControl & Right Hand & 0.299 & 0.692 & \first{9.519} & 8.13 & 1.27 & 5.19 \\
\textbf{Ours} &  & \first{0.137} & \first{0.787} & 9.734 & \first{0.00} & \first{0.00} & \first{1.09} \\
\hline

OmniControl & Left Foot & \first{0.280} & 0.696 & \first{9.553} & 5.94 & 0.94 & 3.14 \\
\textbf{Ours} &  & 0.368 & \first{0.768} & 9.774 & \first{0.00} & \first{0.00} & \first{1.14} \\
\hline

OmniControl & Right Foot & \first{0.319} & 0.701 & \first{9.481} & 6.66 & 1.20 & 3.34 \\
\textbf{Ours} &  & 0.361 & \first{0.775} & 9.778 & \first{0.00} & \first{0.00} & \first{1.16} \\
\hline

OmniControl & All Joints above & 2.614 & 0.606 & 8.594 & 75.59 & 12.30 & 23.67 \\
\textbf{Ours} &  & \first{0.032} & \first{0.794} & \first{9.750} & \first{0.00} & \first{0.00} & \first{1.57} \\
\hline
\end{tabular}
}
\caption{Quantitative results of comparison with state-of-the-art methods on Humanml3D test set. The best scores are highlighted in red.}
\label{tab:humanml3d_all}
\end{table}

% avverage line
% masking strategy different
% the order of contribution
\begin{table}[h!]
\centering
\resizebox{0.85\textwidth}{!}{
\begin{tabular}{c|c|c|c|c|c|c|c}
\hline
% Method & Control Joint & FID $\downarrow$ & R-precision $\uparrow$ (Top-3) & Diversity $\rightarrow$ & Traj. Err. $\downarrow$ (50 cm, \%) & Loc. Err. $\downarrow$ (50 cm, \%) & Avg. Err. (cm) $\downarrow$ \\
Method & Control Joint & FID↓ & \makecell[c]{R-precision↑\\(Top-3)} & Diversity→ & \makecell[c]{Traj. Err.↓\\(50 cm, \%)} & \makecell[c]{Loc. Err.↓\\(50 cm, \%)} & \makecell[c]{Avg. Err.\\(cm)↓} \\
\hline
Real & - & 0.031 & 0.779 & 11.08 & 0.000 & 0.000 & 0.000 \\ \hline
PriorMDM & & 0.851 & 0.397 & 10.518 & 33.10 & 14.00 & 23.05 \\
GMD & Pelvis & 1.565 & 0.382 & 9.664 & 54.43 & 30.03 & 40.70 \\
OmiControl & & 0.702 & 0.397 & \first{10.927} & 11.05 & 3.37 & 7.59 \\
\textbf{Ours} & & \first{0.432} & \first{0.757} & 10.723 & \first{0.28} & \first{0.11} & \first{2.76} \\ 

\hline
OmiControl & Average & 0.788 & 0.379 & \first{10.841} & 14.33 & 3.68 & 8.54 \\
\textbf{Ours} & & \first{0.487} & \first{0.751} & 10.716 & \first{0.52} & \first{0.15} & \first{2.98} \\
\hline
\end{tabular}
}
\caption{Quantitative results of comparison with state-of-the-art methods on KIT test set. The best scores are highlighted in red.}
\label{tab:kit_all}
\end{table}

\subsubsection{Implementation Details}
In the part-based VQ-VAE, the settings of the encoder and the decoder module are based on T2M-GPT \cite{T2M-GPT} with the downsampling $s$ factor set to 4. For the masked trajectory transformer, we use a frozen CLIP-ViT-B/32 \cite{CLIP} model for pre-processing the text prompt $\bm{L}$, and then a 4-layer transformer processes the text information and control trajectories $\bm{R'}$ into the latent features, followed by a 3-layer transformer decodes the latent features into the codes indices. The embedding dimension of these two stage transformers are 512 and 256 respectively. The networks are trained using AdamW optimizer with learning rate decays from 1e-4 to 1e-7 with a batch size of 64. For the run-time optimization step, we use Limited-Memory BFGS  \cite{LMBFGS} as the optimization algorithm with learning rate set to 0.1. The accuracy criteria is set to 1E-6 and we provide an ablation study of the criteria setting in Sec.~\ref{sec:exp_optimization_scheme}. 

In this experiments section, we use a machine with RTX4090 GPU for running the tests except in Sec.~\ref{sec:exp_runtime} for a fair comparison. The testing batch size is 32 following previous works using HumanML3D \cite{Guo_2022_CVPR_humanml3d} and KIT-ML \cite{kitdataset} dataset.

\subsection{Controllable Human Motion Generation}
\label{sec:exp_motion}

\subsubsection{Qualitative Results}
In Fig.~\ref{fig:qualitative} and Fig.~\ref{fig:qualitative_2}, we showcase some qualitative results of our method using various control trajectories from: 1) Specifying parameters of analytical shapes; 2) Hand drawings in 3D space; and 3) Environment settings. This versatility underlines our method's practical applicability across a wide range of scenarios, demonstrating its significant utility in diverse applications.

\begin{wraptable}{r}{0.438\columnwidth}
\footnotesize
\centering
\resizebox{0.38\columnwidth}{!}{
\begin{tabular}{c|c|c|c|c}
\hline
\makecell[c]{Root Traj \\Masked Rate} & 0\%   & 25\%  & 50\% & 75\% \\ \hline
MModality $\uparrow$      & 0.994 & 1.379 & 1.790 & 1.982 \\ \hline
\end{tabular}
}
\caption{The MModality of the generated motions under different trajectory masking rates.}
\label{table:car_uncond}
\end{wraptable}
\subsubsection{Diverse Output Sampling}
\begin{figure}[t!]
  \centering
  %\vspace{-8mm}
   \includegraphics[width=\linewidth]{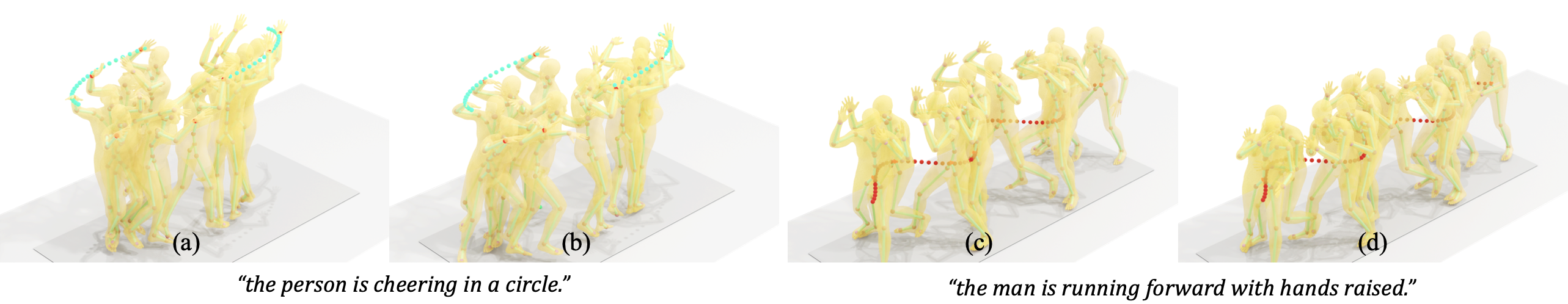}
   %\vspace{-3mm}
   \caption{Qualitative results under the same control trajectories and text prompts.}
   \label{fig:generative}
   %\vspace{-11mm}
\end{figure}

Our approach enables the sampling of multiple outputs for the same trajectory and language input, as illustrated in Figure~\ref{fig:generative}. We conduct an experiment on the test set of HumanML3D, showing the sampling capability of our method by using text and root trajectory in Tab.~\ref{table:car_uncond}.

\begin{figure}[!b]
  \centering
  %\vspace{-6mm}
  \begin{minipage}[!t]{0.49\textwidth}
  \centering
    \includegraphics[width=0.85\linewidth]{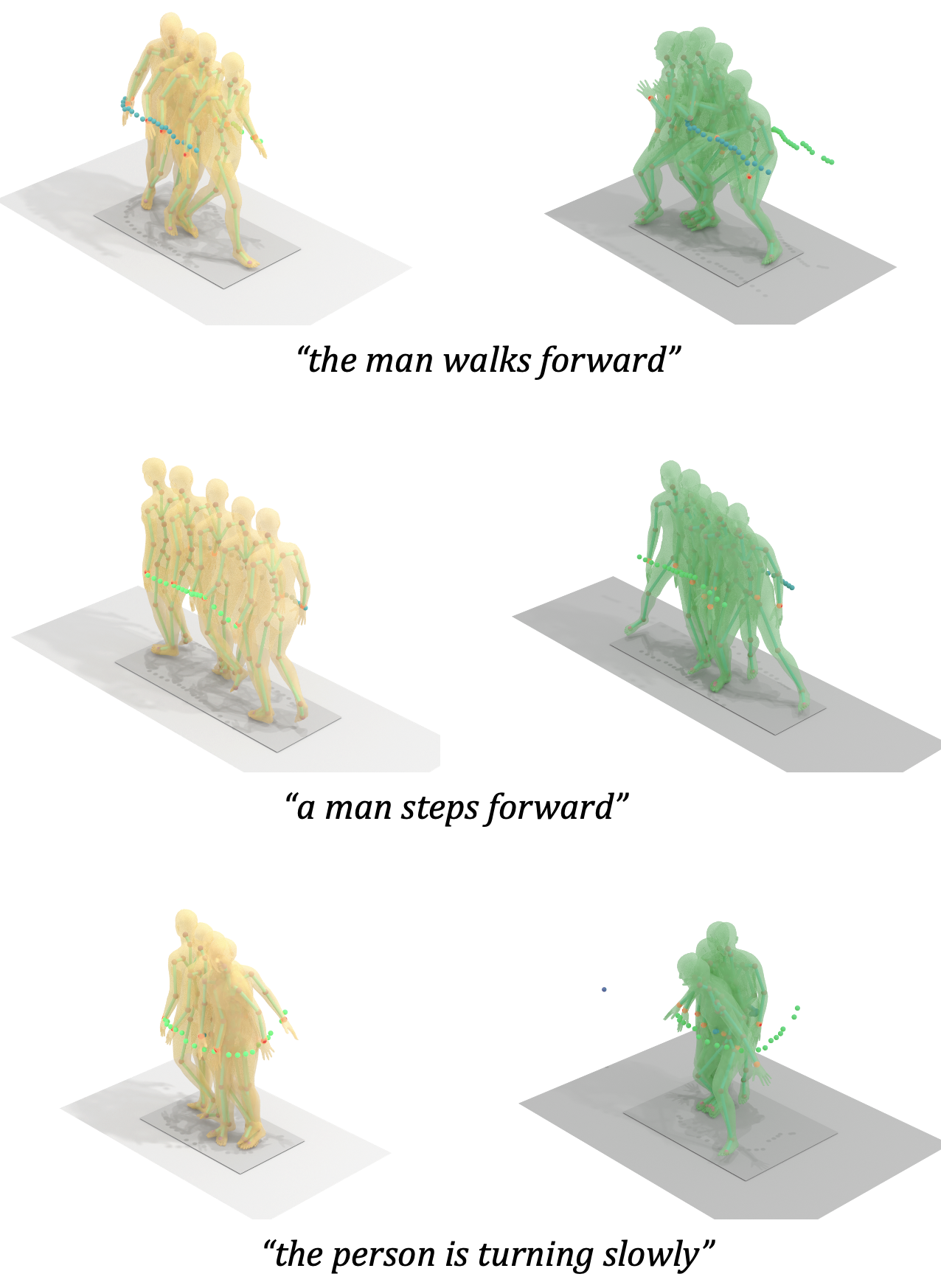}
    %\vspace{-7mm}
    \caption{Qualitative comparison with Omnicontrol\cite{omnicontrol} using the samples from the test set of HumanML3D \cite{Guo_2022_CVPR_humanml3d}. Our results are in \textcolor{yellow}{Yellow} and Omnicontrol's in \textcolor{green}{Green}.
    }
    \label{fig:comparison1}
  \end{minipage}
  \hfill
  \begin{minipage}[!t]{0.49\textwidth}
  \centering
    \includegraphics[width=0.85\linewidth]{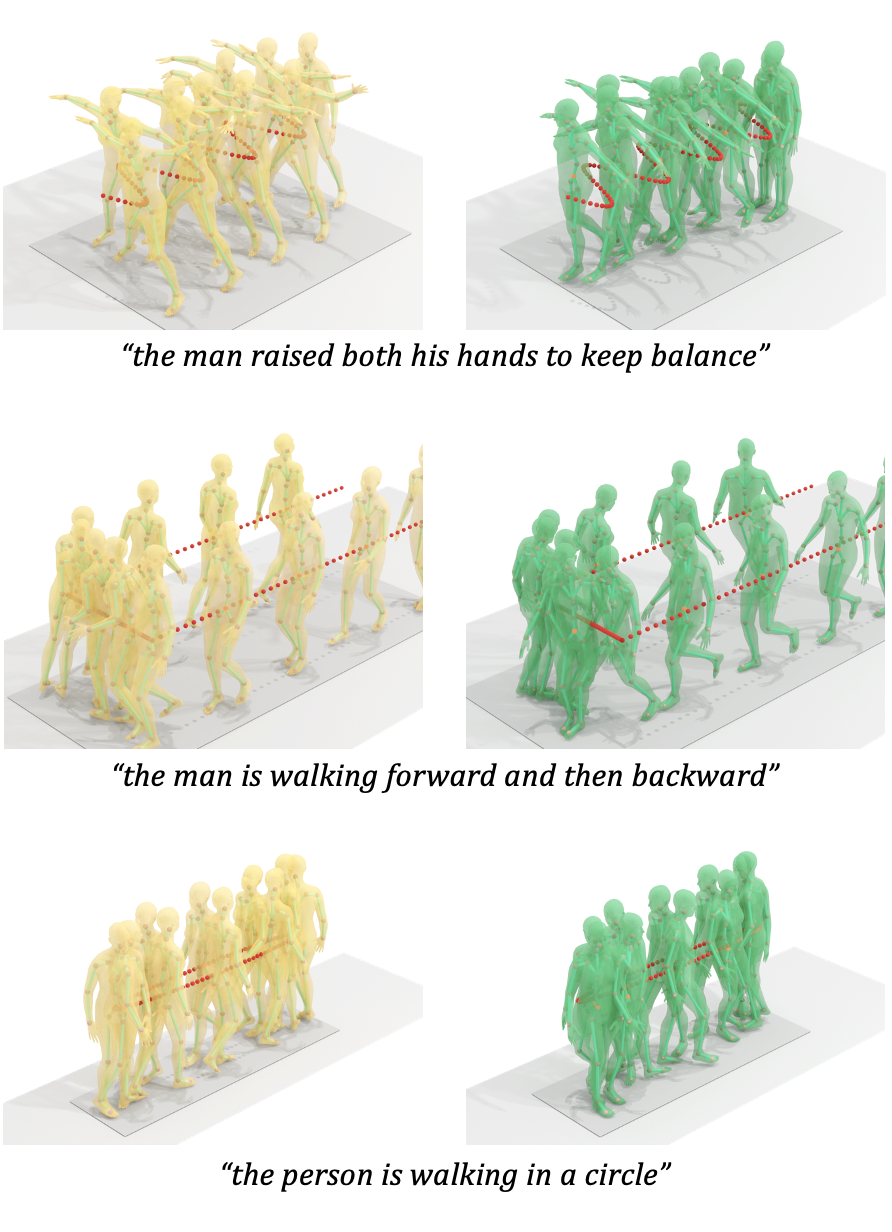}
    \caption{
    Qualitative comparison results with GMD \cite{karunratanakul2023guided} in the task of controlling root path. Our results are in \textcolor{yellow}{Yellow} and GMD's in \textcolor{green}{Green}.
    }
    \label{fig:comparison2}
  \end{minipage}
\end{figure}

\subsubsection{Comparison with SOTAs}
We compare \shortname in controllable motion generation with the SOTA methods. When comparing with MDM~\cite{mdm}, PriorMDM~\cite{newmdm}, and GMD~\cite{karunratanakul2023guided}, we focus on controlling the pelvis only for fair comparisons. We further compare with Omnicontrol~\cite{omnicontrol} \highlight{with more settings of using different end-effectors and the combination of joints.} The results are shown in Tab.~\ref{tab:humanml3d_all} and Tab.~\ref{tab:kit_all}. For the single joint controlling task, our method achieves better scores in R-precision while having comparable results in other fidelity metrics. Meanwhile, our method consistently surpasses other models in the metrics measuring control accuracy. On the HumanML3D, the results generated by our method steadily remained within a 50 cm range of the intended control signals, showcasing remarkable accuracy and stability. We also demonstrate enhanced control accuracy on KIT dataset for using the pelvis only and the average of controlling the joints individually, where \shortname consistently outperforms other methods in accuracy. When using all six joints for controlling, Omnicontrol struggles with multi-joint controls. In contrast, our method can handle multi-joint control effectively, producing highly realistic outcomes with such abundant control signals. Visual comparison is shown in Fig.~\ref{fig:comparison_Omni2}, Fig.~\ref{fig:comparison1} and Fig.~\ref{fig:comparison2}.

\begin{wraptable}{r}{0.332\textwidth}
\centering
    \begin{minipage}[t]{0.31\textwidth} % 使用 [t] 选项顶端对齐
        \centering
        \resizebox{\textwidth}{!}{ % 调整表格的宽度
            \renewcommand{\arraystretch}{0.982} % 减小表格行高
\begin{tabular}{c|c}
  \hline
  \textbf{Methods} & \textbf{Time (s/frame) $\downarrow$} \\ 
  \hline
  OmniControl & 0.606 \\ 
  \hline
  MDM & 0.192 \\ 
  \hline
  GMD & 0.567 \\ 
  \hline
  Ours & \textbf{0.015} \\ 
  \hline

\end{tabular}
}
        \caption{Runtime of different methods.}
        \label{tab:method_runtime_comparison}
    \end{minipage}
\end{wraptable}

\subsection{Run-time Performance}
\label{sec:exp_runtime}
We further compared our method with the SOTA methods in terms of run-time performance. We run the methods on a machine with an RTX3080Ti GPU and record the average process time for a test sample from HumanML3D \cite{Guo_2022_CVPR_humanml3d}. As shown in Tab.~\ref{tab:method_runtime_comparison}, due to our structured and compact latent space and efficient optimization framework, our method consistently outperforms all existing methods by a significant margin.

\begin{figure}[b!]
  %\vspace{-12mm}
  \centering
  % 第一幅图
  \begin{minipage}[t]{0.5\textwidth}
    \centering
     \includegraphics[width=\linewidth]{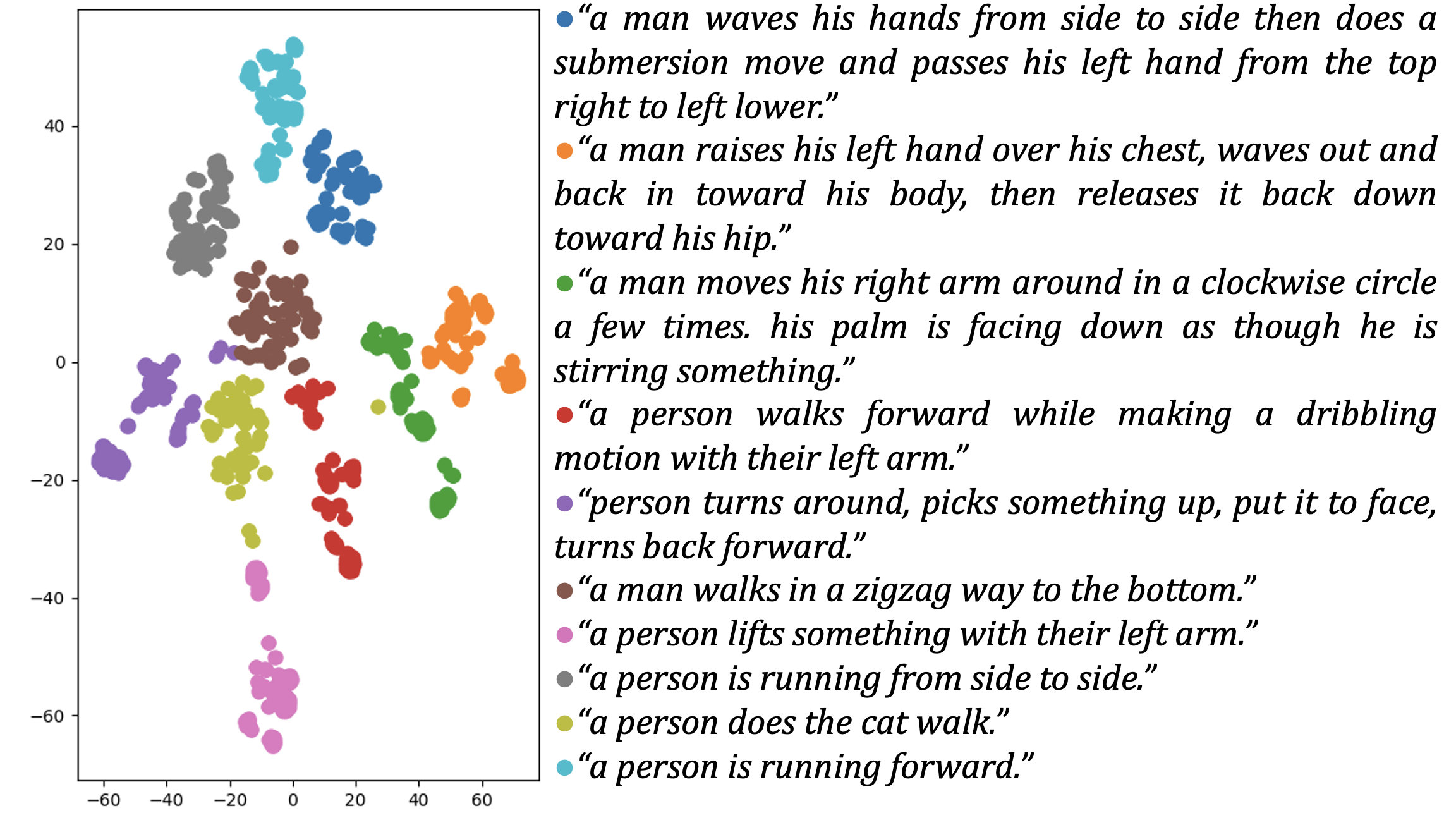}
   %\vspace{-5.5mm}
   \caption{t-SNE visualization of learned latent space. Each color stands for a textural description of the motion. We randomly select ten sentences for visualization.}
   \label{fig:tsne_space}
   %\vspace{-3mm}
  \end{minipage}
  \hfill % 在两幅图之间添加一些空间
  % 第二幅图
  \begin{minipage}[t]{0.48\textwidth}
    \centering
    \includegraphics[width=\linewidth]{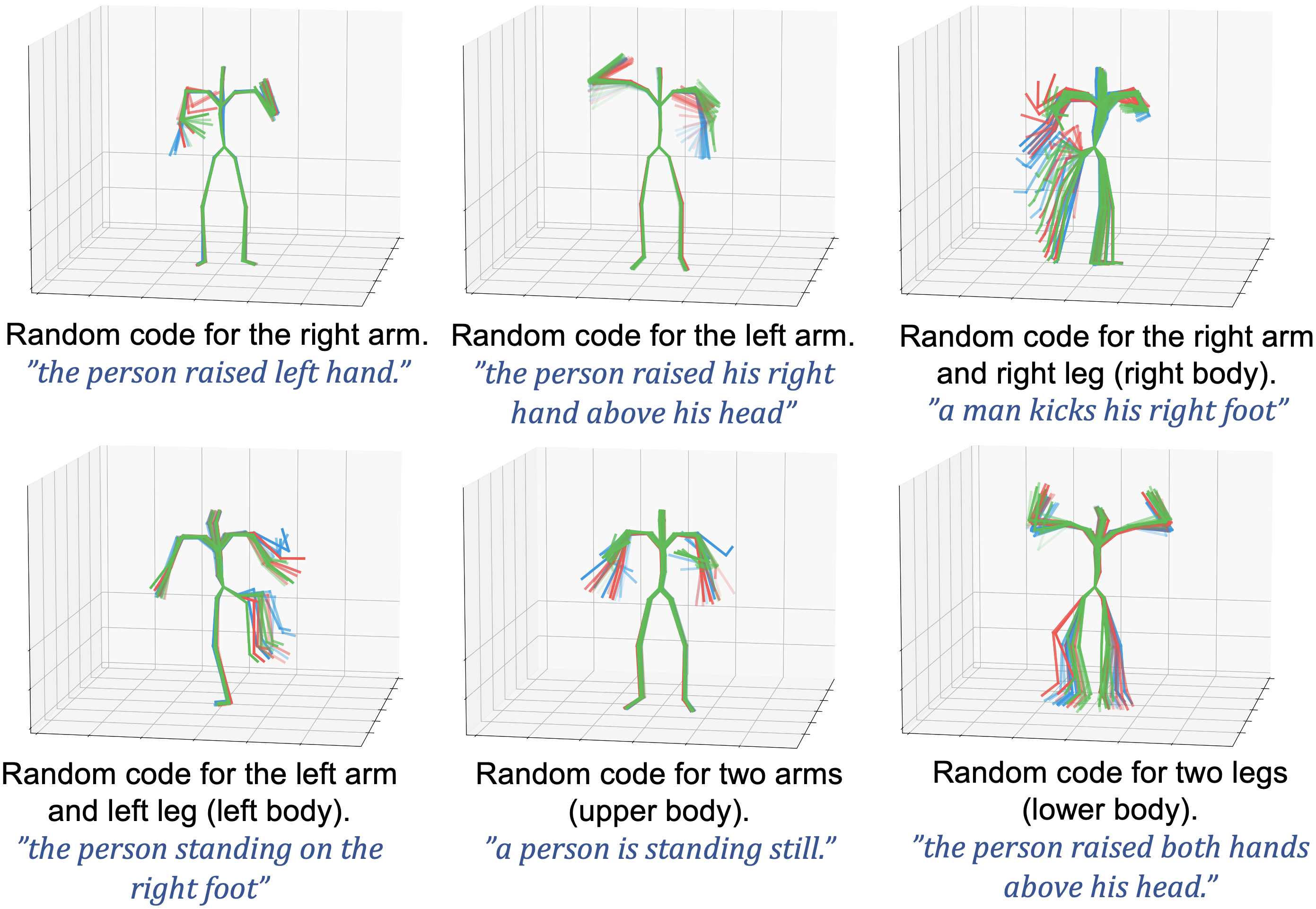} 
   %\vspace{-5.5mm}
   \caption{Motion diversity controlled by language (blue text) and randomly sampled body part latent codes. Three sample motions are in red, blue, and green.}
   \label{fig:latent_variation}
  \end{minipage}
  %\vspace{-8mm}
\end{figure}

\subsection{Ablation Study}

\subsubsection{Motion Latent Space} \label{sec:exp_latent_space}
 We investigate the learned part-based VQ-VAE for motion embedding. In Fig.~\ref{fig:tsne_space}, we randomly choose 10 test cases from HumanML3D test set, and process them with MTT for achieving the corresponding features. We present the result of using t-SNE~\cite{van2008visualizing} for visualizing these features, offering insights into the structure of the acquired motion latent space. Our part-based VQ-VAE contributes a compact motion latent space while preserving the semantics described by language inputs, as evidenced in Fig.~\ref{fig:tsne_space}. Next, we visualize the learned body part-level motion prior. As shown in Fig.~\ref{fig:latent_variation}, our framework for motion embedding learns a well-structured latent space. This results in generating motions that conform to the language description while capturing a diverse distribution of motions under each text condition.

\begin{wraptable}{r}{0.489\textwidth} % 调整此处以控制表格的位置和宽度
\centering
%\vspace{-8mm}
\resizebox{0.43\textwidth}{!}{%
\begin{tabular}{c|ccccc}
  \hline
  Optimize Acc Setting  & 1E-4   & 1E-5 & 1E-6    & 1E-7    & 1E-8     \\ \hline
  Runtime (per batch, s) & 3.48 & 11.89  & 28.86 & 61.43 & 127.90 \\ \hline
  Joint Name            & \multicolumn{5}{c}{Avg. Err (cm)} \\ \hline
  Pelvis                  & 4.92 & 2.03  & 1.20  & 0.71  & 0.45   \\
  Head                    & 5.87 & 2.77  & 1.46  & 0.77  & 0.48   \\
  Left Hand               & 8.41 & 4.14  & 1.83  & 0.89  & 0.56   \\
  Right Hand              & 8.45 & 4.08  & 1.80  & 0.88  & 0.56   \\
  Left Foot               & 6.32 & 3.05  & 1.57  & 0.92  & 0.61   \\
  Right Foot              & 6.37 & 3.04  & 1.55  & 0.91  & 0.60   \\
  \hline
  Avg.                    & 6.72 & 3.19  & 1.57  & 0.85  & 0.54   \\ \hline
\end{tabular}
}
%\vspace{-3mm}
\caption{Average Joint Error v.s. Accuracy criteria during optimization when controlling all the joints.}
%\vspace{-6ex}
    \label{tab:joint_error}
\end{wraptable}
\subsubsection{Optimization Scheme}
\label{sec:exp_optimization_scheme}
Finally, we present the average error in controlling all six joints of our method under varying accuracy criteria during optimization. %\jd{What are the controlled joints here?}
These errors for each joint are recorded in Table~\ref{tab:joint_error}, utilizing the HumanML3D dataset test set. As indicated in Table~\ref{tab:joint_error}, higher optimization accuracy settings result in lower accuracy errors but require more optimization time. In this paper, we consistently set the optimization accuracy to 1E-6.

\section{Conclusion}
%\ZY{pls revise according to the current abs/intro.}
We introduce \shortname, a method for controllable human motion generation using a combination of joint trajectory and language inputs. \shortname leverages a human body morphology-aware structured latent space using a VQ-VAE, and a motion distribution conditioned on language and trajectory using a Masked Trajectory Transformer. We further introduce an efficient optimization framework offers flexible controls by allowing users to specify various optimization goals while ensuring high runtime efficiency. Extensive experiments validate the effectiveness of our framework, highlighting its capability to enable users to quickly generate high-fidelity human motion interactively. \\

%\section{Acknowledgements}
%The authors would like to thank Yiming Xie for preparing sample cases in the comparisons. Dinesh Jayaraman was supported by NSF CAREER Award 2239301.
% \noindent \textbf{Limitation}

% Our framework effectively learns a compact human body morphology-aware structured latent space using a VQ-VAE, and a motion distribution conditioned on language and trajectory using a Masked Trajectory Transformer. The proposed optimization scheme offers flexible controls by allowing users to specify various optimization goals while ensuring high efficiency.}

%\item \ZY{\shortname demonstrates its superior
%control accuracy in producing realistic human motion with remarkable runtime efficiency, compared with SOTAs. Our codes will be publicly available for research purpose upon the paper publication.

%\section{Limitations and Future Works}
%\label{sec:limitations}
% \noindent\textit{Limitations and Future Works} 
% \highlight{In pathological cases where there exist conflicts between input trajectories and languages, we have observed the final output after optimization is largely consistent with the trajectory. This is because TLControl uses languages specifically when generating the coarse initial motion guess, after which it optimizes the motion to match the trajectories. Future enhancements could focus on conflict detection and intuitive user guidance for adjustments, aiming for more accurately aligned animations with user expectations.
% }

\highlight{For the cases where there exist conflicts between input trajectories and languages, generating motions under these conflicting inputs is an open question. Handling such cases is beyond the scope of this paper. In this paper, we assume that the language and trajectory controls do not conflict. Nevertheless, we acknowledge that this issue presents a compelling avenue for future research. Future enhancements could focus on conflict detection and intuitive user guidance for adjustments for more accurately aligned animations with user expectations.  
}

%However, output quality is highest when the user provides consistent $L$ and $R'$.
% \ZY{language traj conflicts: 3. Revise the paper: Clarify the impact of trajectory lines and verbal descriptions on the final animation, especially when the two are in conflict.}
% \ZY{traj conflicts.}

% double checking: \url{https://arxiv.org/pdf/2310.08580.pdf}

\clearpage

\section*{Acknowledgment}
This work is partly supported by the Innovation and Technology Commission of the HKSAR Government under the ITSP-Platform grant (Ref: ITS/319/21FP) and the InnoHK initiative (TransGP project). The authors would like to thank Yiming Xie for preparing sample cases in the comparisons. Dinesh Jayaraman was supported by NSF CAREER Award 2239301.

\bibliographystyle{splncs04}
\bibliography{main}

\clearpage
\begin{center}
    \textbf{\large TLControl: Trajectory and Language Control for Human Motion Synthesis
\\
\bigskip
Supplementary Material}
\end{center}
\setcounter{section}{0}
\setcounter{figure}{0}
\setcounter{table}{0}

\appendix

This supplementary material covers: the ablation study of our part-based VQ-VQE embedding (Sec.~\ref{supp:vae_embedding}); an experiments of our multi-joint control results (Sec.~\ref{supp:multi-joint}); the performance of our method using incompleted trajecotries (Sec.~\ref{supp:traj_incomp}); the runtime of different joint control strategy (Sec.~\ref{supp:joint_strategy_time}); an ablation study with IK based solution (Sec.~\ref{supp:IK_ablation}); results of using only one modality input (Sec.~\ref{supp:one_modality}); more implementation details (Sec.~\ref{supp:implementation});  and more qualitative results~(Sec.~\ref{supp:qualitative});. 

We also include a video showing more visual results in our supplementary. We highly encourage our readers to view the
supplementary video for our qualitative comparison and results.

\section{Ablation Study on Part-based VQ-VAE}
\label{supp:vae_embedding}
\def\otherEmb{Unsplit VQ-VAE\xspace}

\begin{table*}[h]
%\vspace{-2mm}
\centering
\resizebox{0.99\textwidth}{!}{
\begin{tabular}{|c|c|c|c|c|c|c|}
\hline
Method & Control Joint & FID↓ & R-precision↑ (Top-3) & Diversity→ & Avg. Err. (cm)↓ & Runtime Per Batch (s)↓ \\
\hline
Real & - & 0.002 & 0.797 & 9.503 & 0.00 & - \\
\hline
\otherEmb & Pelvis & 0.334 & 0.767 & 9.681 & 1.52 & 22.81 \\
\textbf{Ours} &  & \textbf{0.271} &\textbf{0.779} & \textbf{9.569} & \textbf{1.08} & \textbf{17.67} \\
\hline
\otherEmb & Head & 0.325 & 0.768 & 9.687 & 1.61 & 26.61 \\
\textbf{Ours} &  & \textbf{0.279} & \textbf{0.778} & \textbf{9.606} & \textbf{1.10} & \textbf{19.80} \\
\hline
\otherEmb & Left Hand & 0.156 & 0.786 & 9.776 & 1.57 & 30.22 \\
\textbf{Ours} &  & \textbf{0.135} & \textbf{0.789} & \textbf{9.757} & \textbf{1.08} & \textbf{25.35} \\
\hline
\otherEmb & Right Hand & 0.145 & 0.783 & 9.793 & 1.60 & 29.89 \\
\textbf{Ours} &  & \textbf{0.137} & \textbf{0.787} & \textbf{9.734} & \textbf{1.09} & \textbf{25.17} \\
\hline
\otherEmb & Left Foot & 0.433 & 0.753 & 9.839 & 1.59 & 25.29 \\
\textbf{Ours} &  & \textbf{0.368} & \textbf{0.768} & \textbf{9.774} & \textbf{1.14} & \textbf{20.90} \\
\hline
\otherEmb & Right Foot & 0.412 & 0.756 & 9.802 & 1.64 & 25.34 \\
\textbf{Ours} &  & \textbf{0.361} & \textbf{0.775} & \textbf{9.778} & \textbf{1.16} & \textbf{21.15} \\
\hline
\otherEmb & All Joints above & 0.066 & 0.790 & 9.780 & 2.51 & 38.84 \\
\textbf{Ours} &  & \textbf{{0.032}} & \textbf{{0.794}} & \textbf{{9.750}} & \textbf{{1.57}} & \textbf{28.86} \\
\hline
\end{tabular}
}
\caption{Quantitative results of comparison with different embedding design on Humanml3D test set. The best results are highlighted in red.}
\label{sup_tab:vq_embeding}
%\vspace{-4mm}
\end{table*}

To validate the effectiveness of the proposed part-based VQ-VAE, we compared our network structure with a counterpart VQ-VAE without body-part level disentangling, which we referred to as \textbf{Unsplit VQ-VAE}. To guarantee a fair comparison, both models were configured with an identical feature dimension that represents the full-body motion and is utilized for the runtime optimization. As summarized in Table \ref{sup_tab:vq_embeding}, our proposed part-based VQ-VAE demonstrated superior performance in generating controllable motion, showing its efficacy in preserving language semantics alongside facilitating flexible trajectory control. This is attributed to our disentanglement design of our VQ-VAE that allows an efficient representation of motion dynamics at the body-part level. Moreover, our part-based VQ-VAE exhibits the capability to encode motions into a more compact latent space. This compactness not only enhances the model's accuracy in handling complex motion but also significantly reduces processing time during the runtime optimization, making it more suitable for real-time applications where rapid motion generation and adaptation are crucial. 

\begin{figure}[h]
  \centering
  %\vspace{3mm}
  \includegraphics[width=0.7\textwidth]{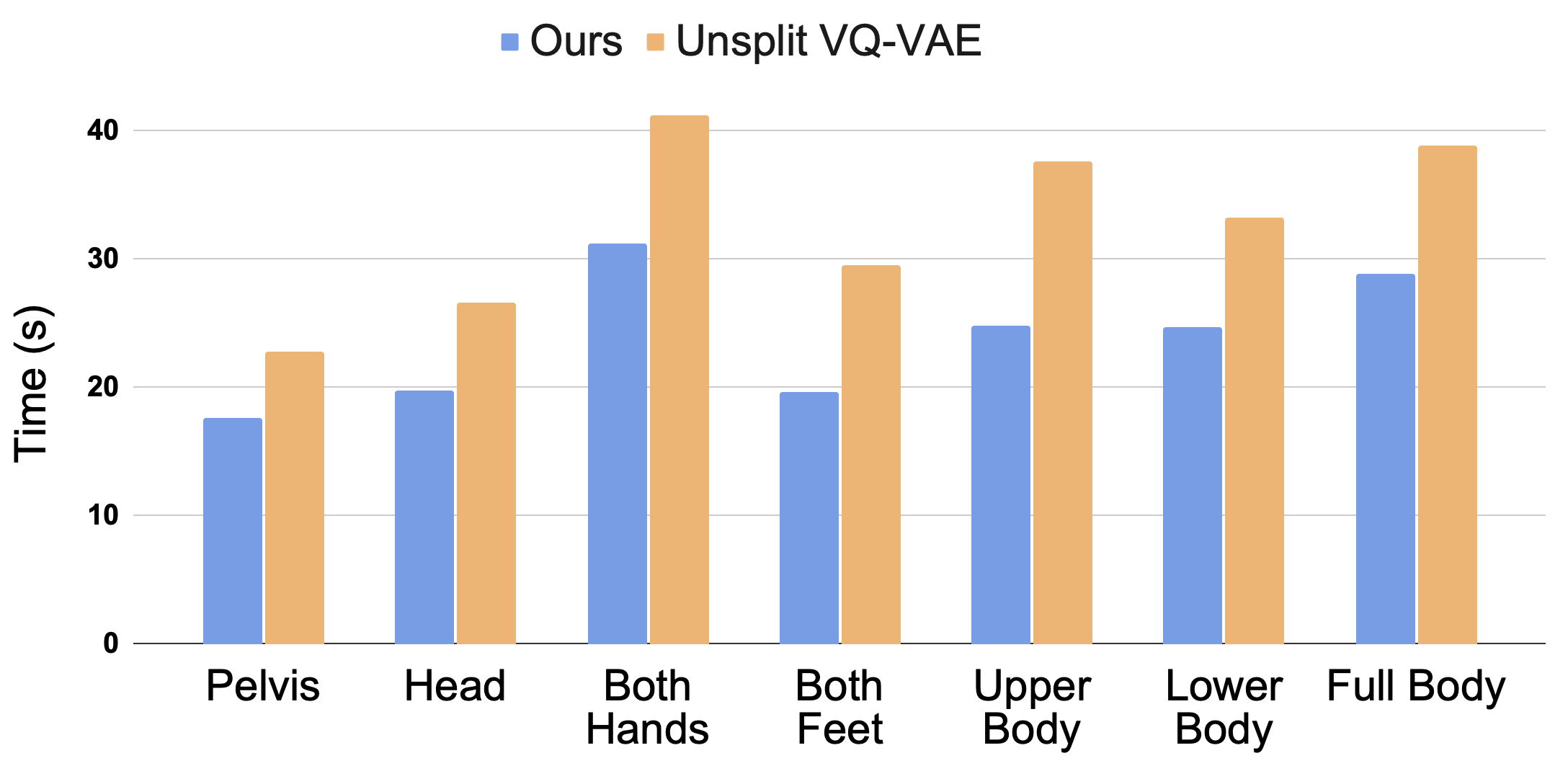}
  \caption{Per batch running time statistics of our embedding comparing to the unsplit embedding. “Upper Body” includes the joints of the hands and the head joint. “Lower Body” includes the joints of two feet and the joint of the pelvis.} %\ZY{Could remove the "bold" effects to make it consistent with the main paper?}}
  %\vspace{1mm}
  \label{fig:joint_time_unsectioned}
\end{figure}

Figure \ref{fig:joint_time_unsectioned} illustrates the comparative analysis of the average processing times for motion synthesis using our part-based embedding versus the unsplit embedding under various control strategies. On average, our embedding design demonstrates a significant reduction in processing time, specifically a 27.3\% decrease compared to the unsplit VQ-VAE embedding. This efficiency gain is primarily attributed to the optimized structure of our part-based embedding, which streamlines the synthesis process by effectively handling the complexities of motion data.

%\begin{figure}%[H]
   % \vspace{-2mm}
%  \centering
%  \vspace{2mm}
%   \includegraphics[width=\linewidth]{imgs/visual_compare_GMD.jpg}   
   %\vspace{-4mm}
%   \caption{Qualitative comparison results with GMD \cite{karunratanakul2023guided} in the task of controlling root path. Our results are shown in \textcolor{yellow}{Yellow}, while the results of GMD are depicted in \textcolor{green}{Green}. Please refer to our supplementary video for more details of the comparison. }
%   \label{fig:comparison_GMD}
%    \vspace{2mm}
%\end{figure}

\section{Multi-Joint Control Results}
\label{supp:multi-joint}

\begin{table}[h]
%\vspace{-4mm}
\centering
\resizebox{0.8\textwidth}{!}{
\begin{tabular}{|c|c|c|c|c|}
\hline
Control Joint & FID $\downarrow$ & R-precision $\uparrow$ (Top-3) & Diversity $\rightarrow$ & Avg. Err. (cm) $\downarrow$ \\
\hline
%- (Real) & 0.031 & 0.779 & 11.08 & 0.000 \\ \hline
Both Hands & 0.108 & 0.789 & 9.747 & 1.25 \\
Upper Body & 0.088 & 0.791 & 9.740 & 1.35 \\
Both Feet & 0.320 & 0.775 & 9.748 & 1.29 \\
Lower Body & 0.249 & 0.777 & 9.746 & 1.38 \\
\hline
\end{tabular}
}
\caption{Quantitative results of multi-joint control. “Upper Body” includes the joints of hands and the head joint. “Lower Body” includes the joints of two feet and the joint of pelvis.}
%\vspace{-4mm}
\label{sup_tab:multi_joint}
\end{table}

In this section, we present the performance metrics of our method when applied to various groups of joints. This test is performed in the test set of HumanML3D dataset \cite{Guo_2022_CVPR_humanml3d}, and the results are shown in Table ~\ref{sup_tab:multi_joint}. Our approach demonstrates notable efficiency and accuracy in managing multiple joints simultaneously. As we integrate additional control trajectories, the motion representation becomes increasingly detailed. This enhancement allows our method to generate more realistic results, showcasing its adaptability and precision in complex joint operations.

\section{Trajectory Incompleteness}
\label{supp:traj_incomp}
\begin{figure}[htbp]
  \centering
  \includegraphics[width=0.45\textwidth]{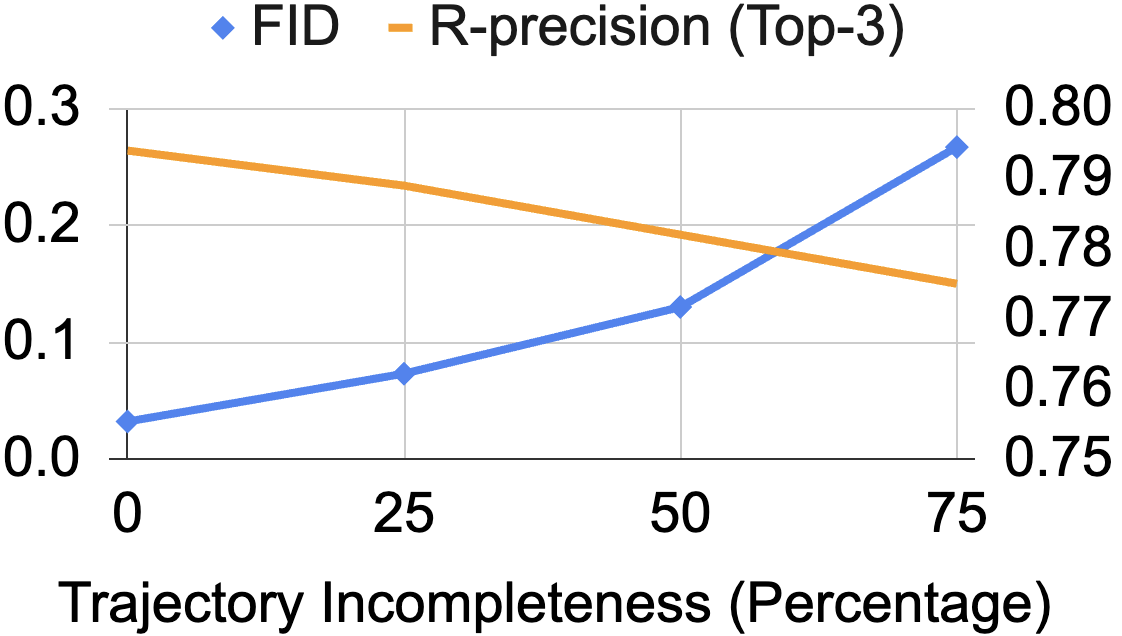}
  \caption{Influence of different trajectory incompleteness. We simulate the incompleteness by applying random masking. The left vertical axis represents the FID metric, while the right vertical axis indicates the R-precision metric.}
  \label{fig:masking}
\end{figure}

We investigate the performance of our method under different levels of trajectory incompleteness, for which we simulate various incompleteness scenarios by adjusting the ratio of random masking. As revealed in Figure ~\ref{fig:masking}, our model's performance degrades gracefully even as the control trajectories become more incomplete. 
As a reference, note that even with full joint trajectories, the performance of previous methods~\cite{omnicontrol} in terms of FID and R-precision are $2.614$ and $0.606$. %\jd{I can't tell on Fig 6 which y-axis is which. Please mark.}

\begin{figure}[]
  \centering
  \includegraphics[width=0.45\textwidth]{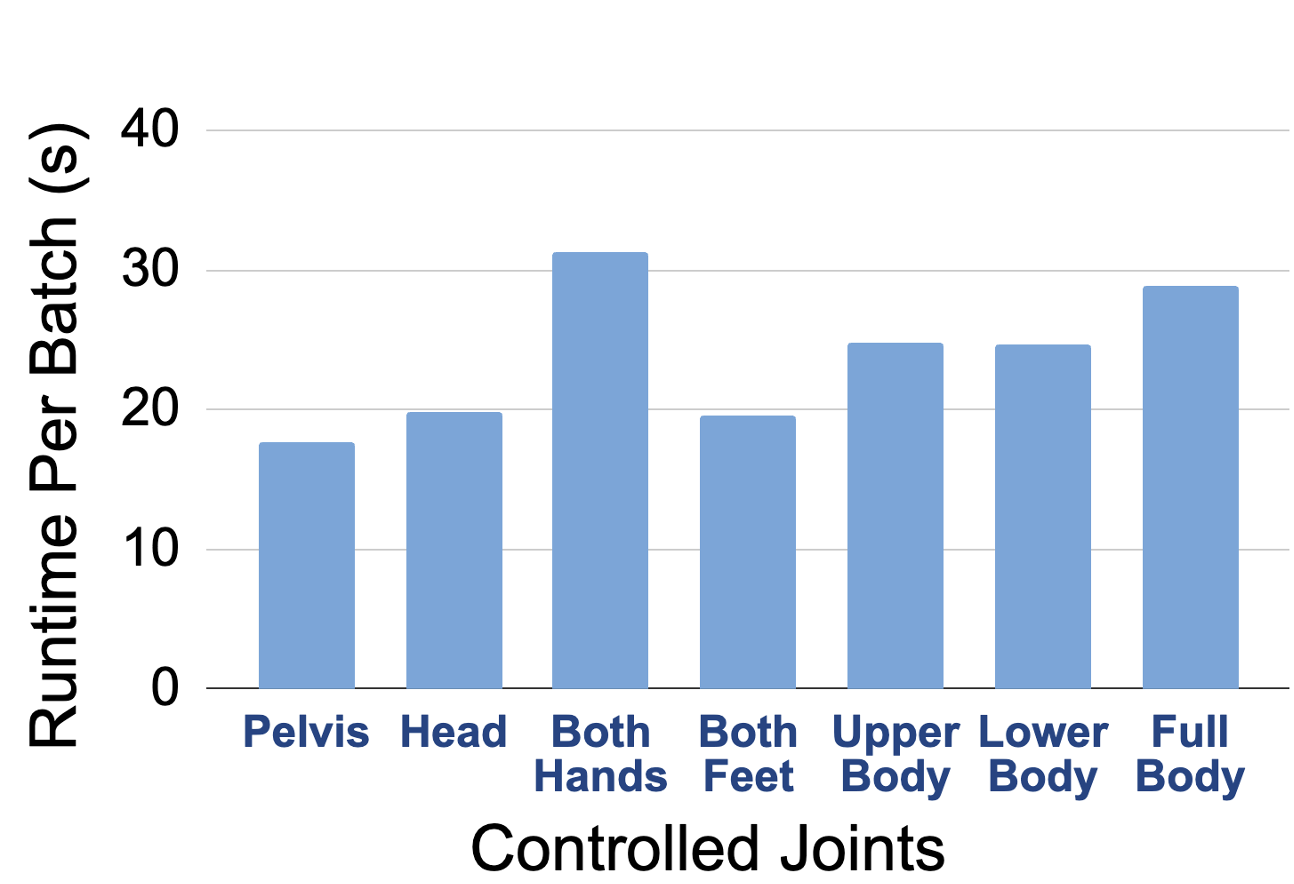}
  \caption{Running time statistics of our optimization when applying different controlled joints. “Upper Body” includes the joints of the hands and the head joint. “Lower Body” includes the joints of two feet and the joint of the pelvis.}
  \label{fig:joint_time}
\end{figure}

\section{Runtime of Different Joint Control Strategy}
\label{supp:joint_strategy_time}

We present the runtime associated with different joint control strategies. In Figure ~\ref{fig:joint_time}, we record the average runtime for processing a batch under the condition of using different combinations of joint trajectories. Due to the complexity of hand motions, optimization that exclusively using both hands trajectories is the most time-consuming.

\section{Ablation with IK Based Solution}
\label{supp:IK_ablation}
\begin{table*}[h]
%\footnotesize
%\vspace{-4mm}
% \hspace*{-5mm}
\centering
\resizebox{0.7\columnwidth}{!}{
\begin{tabular}{|l|cc|cc|cc|}
\hline
Control Strategy & \multicolumn{2}{c|}{No-opt} & \multicolumn{2}{c|}{Joint-IK} & \multicolumn{2}{c|}{\textbf{Ours (Latent-Opt)}} \\ \hline
Traj Masked Rate   & 0\%          & 50\%         & 0\%           & 50\%          & 0\%                & 50\%              \\ \hline
FID (Root) $\downarrow$      & 0.343        & 0.391        & 0.336         & 0.675         & \textbf{0.271}              & \textbf{0.311}             \\ \hline
FID (LHand) $\downarrow$     & 0.226        & 0.297        & 0.383         & 0.612         & \textbf{0.135}              & \textbf{0.224}             \\ \hline
\end{tabular}
}
\caption{Quantitative results of comparing IK based solution.}
\end{table*}
Furthermore, we conduct a comparative analysis between the use of inverse kinematics (IK) and our optimization framework. In this experiment, we control the root and left hand separately using trajectories with different mask rates. We report the FID for motions: 1) decoded from initial state features without optimization; 2) optimized from the decoded motions over joints using IK; 3) optimized from the decoded motions over the latent features using our design. Spatially, IK is for joint level adjustment, and it is limited in maintaining the semantics of the overall body motion. Temporally, in cases of incomplete control trajectories, IK cannot guide the frames without trajectories, leading to discrepancies between the generated motions and the intended control goals.

\section{Using Only One Modality Input}
\label{supp:one_modality}
\begin{figure*}[h]
%\vspace{-10mm}
  \centering
\includegraphics[width=0.65\linewidth]{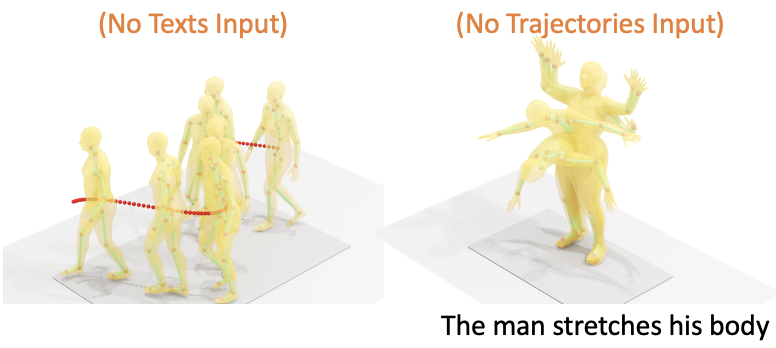}
   \caption{Qualitative results of using only one modality.}
   \label{fig:one_modality}
   %\vspace{-6mm}
\end{figure*}
TLControl is compatible with empty trajectories $R$ or empty language input $L$ for pure language-conditioned or trajectories-conditioned motion generation, although we typically assume that trajectory and language descriptions are both given. Here we give examples of using language only and using trajectories only for motion generation in Figure~\ref{fig:one_modality}, where our method produces high-quality results.

\section{Implementation Details}
\label{supp:implementation}
In this section, we provide a more detailed explanation of the implementation of our method. Following~\cite{T2M-GPT}, our encoder and the decoder consist of standard 1D convolutional layers, residual blocks \cite{he2016deep}, and ReLU activation functions. The settings of each encoder and the decoder of our part-based VQ-VAE are based on T2M-GPT~\cite{T2M-GPT} using standard 1D convolutional layers, residual blocks \cite{he2016deep}, and ReLU activation functions. 

For each joint encoder, we use a codebook of 126 code vectors and each code of dimension 126. Our input data first passes through an 1D convolutional layer, followed by a ReLU activation function. By setting the temporal downsampling rate to 4, the feature is processed by 2 sequences, where each sequence is a compound module that includes a 1D convolutional layer with a stride of 2, which downsamples temporal dimension of the data by a scale of 2, then followed by a residual block. The encoded feature is also normalized by the norm of itself before being quantized. By combining the quantized codes from all 6 joint encoders, we achieve a full-body latent code of size $126 \times 6 = 756$. The decoder has the same structure as the encoder, where the convolution with stride is replaced with nearest interpolation for temporal upsampling, and the full-body pose output dimension is 263 for Humanml3D \cite{Guo_2022_CVPR_humanml3d} dataset and 251 for KIT \cite{kitdataset} dataset. 

In implementing the masked trajectory transformer, we employ a frozen CLIP-ViT-B/32 \cite{CLIP} model for the initial processing of the text prompt, which translates it into a 512-dimensional vector representative of language features. The joint trajectories are grouped with every 4 closest waypoints being bundled into a single temporal token. This bundling is reflective of our choice to use a downsampling rate of 4 within the part-based VQ-VAE framework. Each waypoint group is then mapped onto a 512-dimensional feature space. A standard 4-layer transformer encoder is used to integrate the language and trajectory features into a cohesive latent space, where tokens comprise both the language feature vector and the feature representations of the trajectory at each temporal token. Following this, a 3-layer transformer encoder is designed to transform the latent space into a sequence of logits for determining the code indices. To transform logits into selected code indices, we follow the common idea of applying softmax and then argmax, but in a differentiable manner: we use the Gumbel-Softmax \cite{jang2016categorical} for producing a one-hot decision vector $\mathbf{v}$, followed by selecting the most probable code index using $\mathbf{C}_{k} \cdot \mathbf{v}$ where $\mathbf{C}_{k} \in \mathbb{R}^{d \times |\mathbf{C}|}$ is the codebook for a joint group containing $|\mathbf{C}|$ distinct codes, each of dimension $d$. Then the selected codes are concatenated to serve as the initial full-body state for the VQ-decoder. 
%\ZY{In particular, directly applying argmax leads to non-differentiable problem. In this paper, we use a one hot decision vector to select the code by $xx$ to achieve a "argmax" function in a differentiable manner. }

For the masking strategies, each iteration within a batch has a 50\% probability of using continuous trajectory masking or joint-level masking. The proportion of continuous trajectory masking is gradually increased, starting at 0\% and steadily advancing to 75\%. 

During the run-time optimization stage, the Limited-Memory BFGS \cite{LMBFGS} method is utilized as the optimization technique. The learning rate is configured at 0.1, and we set the precision target at 1E-6. We limit the process to a maximum of 1000 iterations, maintain an update history size of 200, and employ the ‘strong\_wolfe’ condition for the line search function.

\section{More Qualitative Comparisons on Controllable Motion Synthesis}
\label{supp:qualitative}

\begin{figure*}[h]
%\vspace{-10mm}
  \centering
\includegraphics[width=0.65\linewidth]{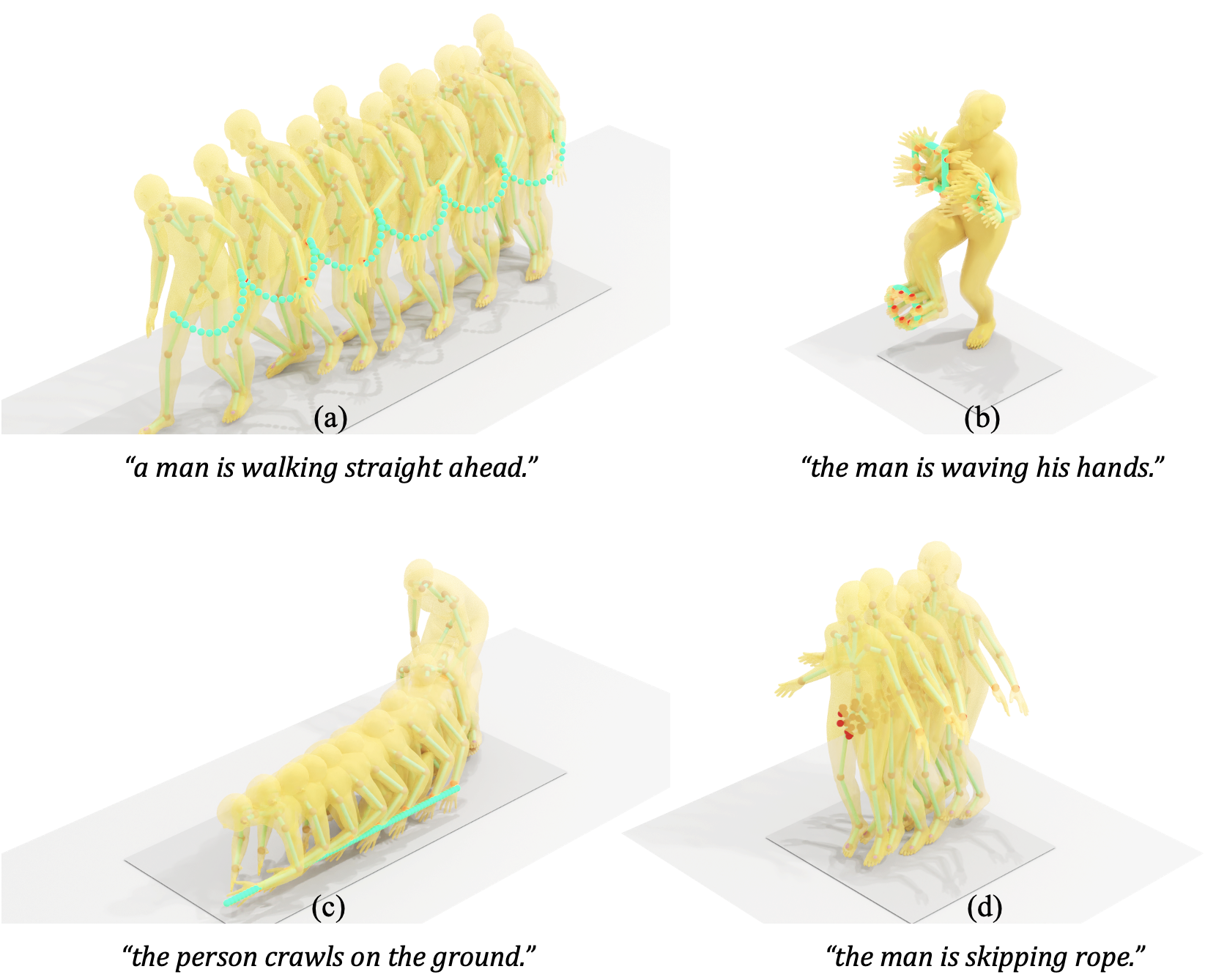}
   \caption{Qualitative results of our method. Please refer to our video for more results.}
   \label{fig:supp_results}
   %\vspace{-6mm}
\end{figure*}

In Figure~\ref{fig:supp_results}, we showcase additional qualitative results. For more demos and comparisons with other methods, please refer to our supplementary video. 

Our method not only demonstrates superior accuracy in adhering to control trajectories and achieves this with a notably reduced runtime but also excels in generating motions that accurately reflect the semantics of text prompts. Our approach exhibits enhanced adaptability and precision in complex motion patterns, outperforming compared methods in dynamic motion scenarios where precise joint control is critical and the ability to capture the essence of textual instructions is paramount.

\end{document}